\documentclass[11pt,a4paper]{article}
\usepackage[hyperref]{acl2021}
\usepackage{times}
\usepackage{latexsym}
\usepackage{latexsym}
\usepackage{xtab,booktabs}
\usepackage[ruled,vlined]{algorithm2e}
\usepackage{amssymb,amsmath,bm}

\usepackage{microtype}
\usepackage{graphicx}
\usepackage{caption}
\usepackage{float}
\usepackage{subcaption}
\usepackage{longtable}
\usepackage{pifont}
\usepackage{CJKutf8}
\usepackage{multirow}
\usepackage{bbm}

\aclfinalcopy 
\def\aclpaperid{1633} 

\newcommand{\cmark}{\ding{51}}
\newcommand{\xmark}{\ding{55}}
\newcommand{\criss}{CRISS}
\newcommand{\simalign}{{\tt SimAlign}}
\newcommand{\fastalign}{{\tt fast\_align}}
\newcommand{\gizapp}{{\tt GIZA++}}
\newcommand{\mbart}{mBART}
\newcommand{\xlmr}{XLM-R}

\title{Bilingual Lexicon Induction \\ via Unsupervised Bitext Construction and Word Alignment}

\author{
   Haoyue Shi
\thanks{\ \ Work done during internship at Facebook AI Research.}
  \\
  TTI-Chicago\\
  \texttt{freda@ttic.edu} \\\And
  Luke Zettlemoyer\\
  University of Washington \\
  Facebook AI Research \\
  \texttt{lsz@fb.com} \\\And
  Sida I. Wang\\
  Facebook AI Research \\
  \texttt{sida@fb.com} 
  }

\date{}

\begin{document}
\begin{CJK*}{UTF8}{bsmi}

\maketitle
\begin{abstract}
Bilingual lexicons map words in one language to their translations in another, and are typically induced by learning linear projections to align monolingual word embedding spaces. In this paper, we show it is possible to produce much higher quality lexicons with methods that combine (1) unsupervised bitext mining and (2) unsupervised word alignment. Directly applying a pipeline that uses recent algorithms for both subproblems significantly improves induced lexicon quality and further gains are possible by learning to filter the resulting lexical entries, with both unsupervised and semi-supervised schemes. Our final model outperforms the state of the art on the BUCC 2020 shared task by 14 $F_1$ points averaged over 12  language pairs, while also providing a more interpretable approach that allows for rich reasoning of word meaning in context. Further analysis of our output and the standard reference lexicons suggests they are of comparable quality, and new benchmarks may be needed to measure further progress on this task.\footnote{Code is publicly available at \url{ https://github.com/facebookresearch/bitext-lexind}.} 
\end{abstract}

\section{Introduction}

\begin{figure*}[t]
\centering
\includegraphics[width=0.8\textwidth]{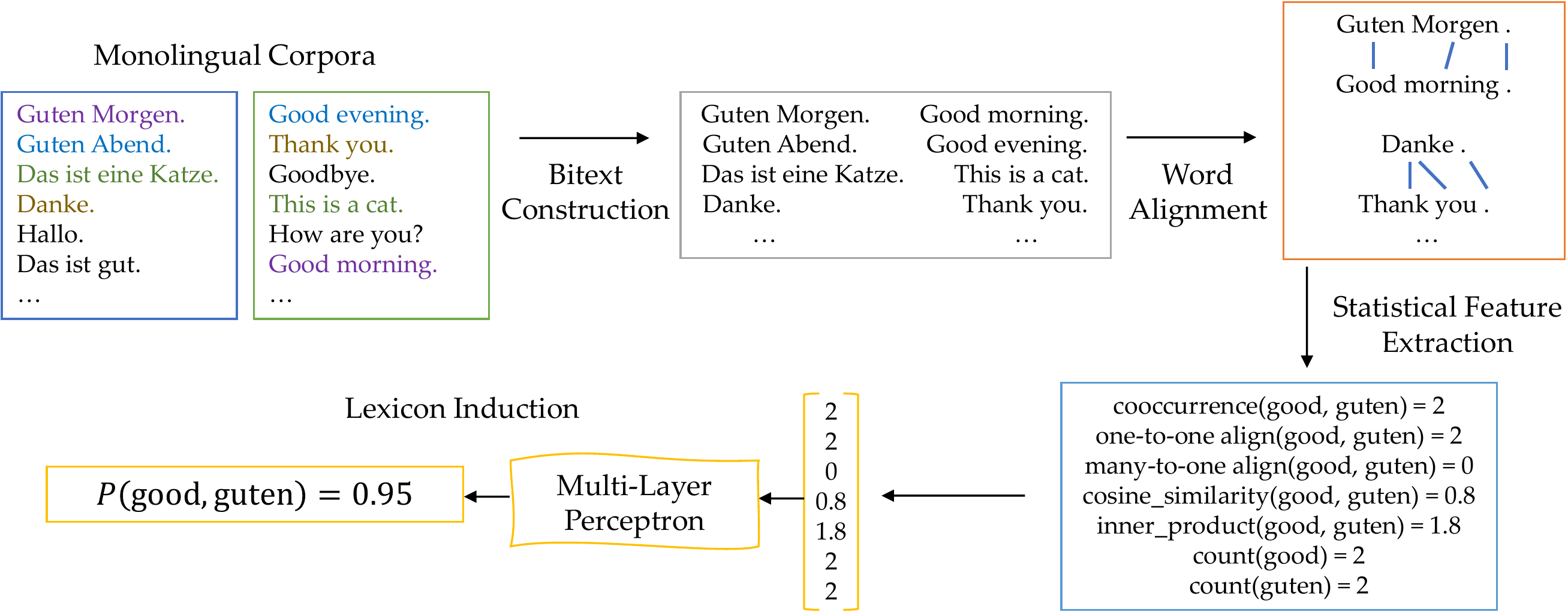}
\caption{\label{fig:teaser} Overview of the proposed retrieval--based supervised BLI framework. Best viewed in color. }
\end{figure*}

Bilingual lexicons map words in one language to their translations in another, and can be automatically induced by learning linear projections to align monolingual word embedding spaces \citep[\emph{inter alia}]{artetxe-etal-2016-learning,smith2017offline,conneau2017word}. Although very successful in practice, the linear nature of these methods encodes unrealistic simplifying assumptions (e.g. all translations of a word have similar embeddings). In this paper, we show it is possible to produce much higher quality lexicons without these restrictions by introducing new methods that combine (1) unsupervised bitext mining and (2) unsupervised word alignment. 


We show that simply pipelining recent algorithms for unsupervised bitext mining~\cite{tran-etal-2020-cross} and unsupervised word alignment~\cite{sabet-etal-2020-simalign} significantly improves bilingual lexicon induction (BLI) quality, and that further gains are possible by learning to filter the resulting lexical entries. Improving on a recent method for doing BLI via unsupervised machine translation~\citep{artetxe-etal-2019-bilingual}, we show that unsupervised mining produces better bitext for lexicon induction than translation, especially for less frequent words.

These core contributions are established by systematic experiments in the class of bitext construction and alignment methods (Figure~\ref{fig:teaser}).
Our full induction algorithm filters the lexicon found via the initial unsupervised pipeline. The filtering can be either fully unsupervised or weakly-supervised: for the former, we filter using simple heuristics and global statistics; for the latter, we train a multi-layer perceptron (MLP) to predict the probability of a word pair being in the lexicon, where the features are global statistics of word alignments.

In addition to BLI, our method can also be directly adapted to improve word alignment and reach competitive or better alignment accuracy than the state of the art on all investigated language pairs.  We find that improved alignment in sentence representations \citep{tran-etal-2020-cross} leads to better contextual word alignments using local similarity \citep{sabet-etal-2020-simalign}.

Our final BLI approach outperforms the previous state of the art on the BUCC 2020 shared task \citep{rapp-etal-2020-overview} by 14 $F_1$ points averaged over 12 language pairs.
Manual analysis shows that most of our false positives are due to the incompleteness of the reference and that
our lexicon is comparable to the reference lexicon and the output of a supervised system.
Because both of our key building blocks make use of the pretrainined contextual representations from mBART \citep{liu2020multilingual} and \criss\ \citep{tran-etal-2020-cross}, we can also interpret these results as clear evidence that lexicon induction benefits from contextualized reasoning at the token level, in strong contrast to nearly all existing methods that learn linear projections on word types. 


\section{Related Work}
\paragraph{Bilingual lexicon induction (BLI).} 
The task of BLI aims to induce a bilingual lexicon (i.e., word translation) from comparable monolingual corpora (e.g., Wikipedia in different languages).
Following \citet{mikolov2013exploiting}, 
most methods train a linear projection to align two monolingual embedding spaces.
For supervised BLI, a seed lexicon is used to learn the projection matrix
\citep{artetxe-etal-2016-learning, smith2017offline, joulin2018loss}.
For unsupervised BLI, the projection matrix is typically found by an iterative procedure such as 
adversarial learning \citep{conneau2017word, zhang-etal-2017-adversarial},
or iterative refinement initialized by a statistical heuristics \citep{hoshen-wolf-2018-non, artetxe-etal-2018-robust}.
\citet{artetxe-etal-2019-bilingual} show strong gains over previous works by word aligning bitext generated with unsupervised machine translation. We show that retrieval-based bitext mining and contextual word alignment achieves even better performance.



\paragraph{Word alignment.} 
Word alignment is a fundamental problem in statistical machine translation, of which the goal is to align words that are translations of each in within parallel sentences~\citep{brown-etal-1993-mathematics}. Most methods assume parallel sentences for training data~\citep[\emph{inter alia}]{och-ney-2003-systematic,dyer-etal-2013-simple,peter2017generating}. 
In contrast, \citet{sabet-etal-2020-simalign} propose \simalign, which does not train on parallel sentences but instead aligns words that have the most similar pretrained multilingual representations \citep{devlin-etal-2019-bert,conneau2019unsupervised}. \simalign\ achieves competitive or superior performance than conventional alignment methods despite not using parallel sentences, and provides one of the baseline components for our work. 
We also present a simple yet effective method to improve performance over \simalign~ (Section~\ref{sec:word-alignment}). 

\paragraph{Bitext mining/parallel corpus mining.} 
Bitext mining has been a long studied task~\citep[\emph{inter alia}]{resnik-1999-mining,shi-etal-2006-dom,abdul-rauf-schwenk-2009-use}.
Most methods train neural multilingual encoders on bitext, which are then used with efficent nearest neighbor search to expand the training set \citep[\emph{inter alia}]{espana2017empirical,schwenk-2018-filtering,guo-etal-2018-effective,artetxe-schwenk-2019-margin}. 
Recent work has also shown that unsupervised mining is possible~\citep{tran-etal-2020-cross,keung-etal-2020-unsupervised}. We use \criss\ \citep{tran-etal-2020-cross}\footnote{\url{https://github.com/pytorch/fairseq/tree/master/examples/criss}} as one of our component models. 



\section{Baseline Components}

We build on unsupervised methods for word alignment and bitext construction, as reviewed below.  

\subsection{Unsupervised Word Alignment}
\simalign\ \citep{sabet-etal-2020-simalign} is an unsupervised word aligner based on the similarity of contextualized token embeddings.
Given a pair of parallel sentences, \simalign\ computes embeddings using pretrained multilingual language models such as mBERT and XLM-R, and forms a matrix whose entries are the cosine similarities between every source token vector and every target token vector.

Based on the similarity matrix, the \emph{argmax} algorithm aligns the positions that are the simultaneous column-wise and row-wise maxima.
To increase recall, \citet{sabet-etal-2020-simalign} also propose \emph{itermax}, which applies \emph{argmax} iteratively while excluding previously aligned positions.

\subsection{Unsupervised Bitext Construction}
\label{sec:unsup-bitext-cons}
We consider two methods for bitext construction: unsupervised machine translation \citep[generation;] [Section~\ref{sec:unsup-bitext-cons-gen}]{artetxe-etal-2019-bilingual} and bitext retrieval \citep[retrieval;][Section~\ref{sec:unsup-bitext-cons-rtv}]{tran-etal-2020-cross}. 

\paragraph{Generation}
\label{sec:unsup-bitext-cons-gen}
\citet{artetxe-etal-2019-bilingual} train an unsupervised machine translation model with monolingual corpora, generate bitext with the obtained model, and further use the generated bitext to induce bilingual lexicons. 
We replace their statistical unsupervised translation model with \criss, a recent high quality unsupervised machine translation model which is expected to produce much higher quality bitext (i.e., translations). 
For each sentence in the two monolingual corpora, we generate a translation to the other language using beam search or nucleus sampling \citep{holtzman2019curious}. 


\paragraph{Retrieval}
\label{sec:unsup-bitext-cons-rtv}
\citet{tran-etal-2020-cross} show that the \criss\ encoder module provides as a high-quality sentence encoder for cross-lingual retrieval: they take the average across the contextualized embeddings of tokens as sentence representation, perform nearest neighbor search with FAISS \citep{johnson2019billion},\footnote{\url{https://github.com/facebookresearch/faiss}} and mine bitext using the margin-based max-score method \citep{artetxe-schwenk-2019-margin}.\footnote{We used max-score~\citep{artetxe-schwenk-2019-margin} as it strongly outperforms the other methods they proposed.} 

The score between sentence representations $\mathbf{s}$ and $\mathbf{t}$ is defined by
\begin{align}
    \label{eq:margin-score}
    &\textit{score}(\mathbf{s}, \mathbf{t}) \\
    = & \frac{\cos\left(\mathbf{s}, \mathbf{t}\right)}{\sum_{\mathbf{t}'\in \textit{NN}_k(\mathbf{t})}\frac{\cos(\mathbf{s}, \mathbf{t}')}{2k} + \sum_{\mathbf{s}'\in \textit{NN}_k(\mathbf{s})}\frac{\cos(\mathbf{s}', \mathbf{t})}{2k}}, \nonumber
\end{align}
where $\textit{NN}_k(\cdot)$ denotes the set of $k$ nearest neighbors of a vector in the corresponding space. 
In this work, we keep the top 20\% of the sentence pairs with scores larger than 1 as the constructed bitext. 

\section{Proposed Framework for BLI}
\label{sec:lexicon-inducer}
Our framework for bilingual lexicon induction takes separate monolingual corpora and the pretrained \criss\ model as input,
and outputs a list of bilingual word pairs as the induced lexicon. 
The framework consists of two parts: (i) an unsupervised bitext construction module which generates or retrieves bitext from separate monolingual corpora without explicit supervision (Section~\ref{sec:unsup-bitext-cons}), and (ii) a lexicon induction module which induces bilingual lexicon from the constructed bitext based on the statistics of cross-lingual word alignment.
For the lexicon induction module, we compare two approaches: fully unsupervised induction (Section~\ref{sec:fully-unsup-lexind}) which does not use any extra supervision, and weakly supervised induction (Section~\ref{sec:weakly-sup-lexind}) that uses a seed lexicon as input. 


\subsection{Fully Unsupervised Induction}
\label{sec:fully-unsup-lexind}
\newcommand{\coc}{\operatorname{coc}}
\newcommand{\mat}{\operatorname{mat}}

We align the constructed bitext with \criss{}-based \simalign, and propose to use smoothed matched ratio for a pair of bilingual word type $\langle s,t \rangle$
\begin{align*}
    \rho(s, t) = \frac{\mat(s,t)}{\coc(s,t) + \lambda}
\end{align*}
as the metric to induce lexicon, where $\mat(s,t)$ and $\coc(s,t)$ denote the one-to-one matching count (e.g., guten-good; Figure~\ref{fig:teaser}) and co-occurrence count of $\langle s,t\rangle$ appearing in a sentence pair respectively, and $\lambda$ is a non-negative smoothing term.%
\footnote{We use $\lambda=20$. This reduces the effect of noisy alignment: the most extreme case is that both $\mat(s,t)$ and $\coc(s,t)$ are $1$, but it is probably not desirable despite the high matched ratio of 1.} 

During inference, we predict the target word $t$ with the highest $\rho(s, t)$ for each source word $s$. 
Like most previous work \citep[][\emph{inter alia}]{artetxe-etal-2016-learning,smith2017offline,conneau2017word}, this method translates each source word to exactly one target word. 

\subsection{Weakly Supervised Induction}
\label{sec:weakly-sup-lexind}
We also propose a weakly supervised method, which assumes access to a seed lexicon. This lexicon is used to train a classifier to further filter the potential lexical entries. 

For a pair of word type $\langle s, t \rangle$, our classifier uses the following global features: 
\begin{itemize}
    \item Count of alignment: we consider both one-to-one alignment (Section~\ref{sec:fully-unsup-lexind}) and many-to-one alignment (e.g., danke-you and danke-thank; Figure~\ref{fig:teaser}) of $s$ and $t$ separately as two features, since the task of lexicon induction is arguably biased toward one-to-one alignment. 
    \item Count of co-occurrence used in Section~\ref{sec:fully-unsup-lexind}.
    \item The count of  $s$ in the source language and $t$ in the target language.%
    \footnote{\simalign\ sometimes mistakenly align rare words to punctuation, and such features can help exclude such pairs. }
    \item Non-contextualized word similarity: we feed the word type itself into \criss, use the average pooling of the output subword embeddings, and consider both cosine similarity and dot-product similarity as features.
\end{itemize} 

For a counting feature $c$, we take $\log \left(c + \theta_c\right)$, where $\theta$ consists of learnable parameters.
There are 7 features in total, which is denoted by $\mathbf{x}_{\langle s,t\rangle} \in \mathbb{R}^7$.

We compute the probability of a pair of words $\langle s, t \rangle$ being in the induced lexicon $P_\Theta(s, t)$\footnote{Not to be confused with joint probability.} by a ReLU activated multi-layer perceptron (MLP): 
\begin{align*}
    \bm{\hat{h}}_{\langle s,t\rangle} &= \mathrm{ReLU}\left(\mathbf{W}_1\mathbf{x}_{\langle s,t\rangle} + \mathbf{b}_1\right) \\
    P_\Theta(s, t) &= \sigma \left(\mathbf{w}_2 \cdot \bm{\hat{h}}_{\langle s,t\rangle} + b_2 \right), 
\end{align*}
where $\sigma(\cdot)$ denotes the sigmoid function, and $\Theta = \{\mathbf{W}_1, \mathbf{b}_1, \mathbf{w_2} ,b_2\}$ denotes the learnable parameters of the model.

Recall that we are able to access a seed lexicon, which consists of pairs of word translations. 
In the training stage, we seek to maximize the log likelihood:
\begin{align*}
    \Theta^* = \arg\max_{\Theta} & \sum_{\langle s,t\rangle \in \mathcal{D}_+} \log P_\Theta( s,t) \\
    + &\sum_{\langle s',t'\rangle \in \mathcal{D}_-}\log \left(1- P_\Theta( s',t')\right),
\end{align*}
where $\mathcal{D}_+$ and $\mathcal{D}_-$ denotes the positive training set (i.e., the seed lexicon) and the negative training set respectively. 
We construct the negative training set by extracting all bilingual word pairs that co-occurred but are not in the seed word pairs. 

We tune two hyperparameters $\delta$ and $n$ to maximize the $F_1$ score on the seed lexicon and use them for inference, where $\delta$ denotes the prediction threshold and $n$ denotes the maximum number of translations for each source word, following \citet{laville-etal-2020-taln} who estimate these hyperparameters based on heuristics. 
The inference algorithm is summarized in Algorithm~\ref{algo:infer-weakly-sup-lexind}.

\begin{algorithm}[t]
\SetAlgoLined
\SetKwInOut{Input}{Output}
\KwIn{Thresholds $\delta, n$,\\ ~~~~~~~~~~~~Model parameters $\Theta$, source words $S$}
\KwOut{Induced lexicon $\mathcal{L}$}
 $\mathcal{L}\leftarrow \emptyset$ \\
 \For{$s \in S$} {
  $ \left(\langle s, t_1 \rangle, \ldots, \langle s, t_k \rangle \right) \leftarrow $ bilingual word pairs sorted by the descending order of $P_\Theta(s, t_i)$ \\ 
  $k' = \max \{j \mid P_\Theta(s, t_j) \geq \delta, j\in[k]\} $ \\ 
  $m = \min(n, k')$ \\
  $\mathcal{L} \leftarrow \mathcal{L} \cup \{\langle s, t_1\rangle, \ldots, \langle s, t_m \rangle\}$
 }
 \caption{\label{algo:infer-weakly-sup-lexind} Inference algorithm for weakly-supervised lexicon induction. }
\end{algorithm}

\section{Extension to Word Alignment}
\label{sec:word-alignment}
The idea of using an MLP to induce lexicon with weak supervision (Section~\ref{sec:weakly-sup-lexind}) can be directly extended to word alignment. 
Let $\mathcal{B} = \{\langle \mathcal{S}_i, \mathcal{T}_i \rangle \}_{i=1}^N$ denote the constructed bitext in Section~\ref{sec:unsup-bitext-cons}, where $N$ denotes the number of sentence pairs, and $\mathcal{S}_i$ and $\mathcal{T}_i$ denote a pair of sentences in the source and target language respectively. 
In a pair of bitext $\langle \mathcal{S}, \mathcal{T}\rangle$, $\mathcal{S} = \langle s_1, \ldots, s_{\ell_s} \rangle$ and $\mathcal{T} = \langle t_1, \ldots, t_{\ell_s} \rangle$ denote sentences consist of word tokens $s_i$ or $t_i$.

For a pair of bitext, \simalign\ with a specified inference algorithm produces word alignment $\mathcal{A} = \{\langle a_i, b_i \rangle\}_i$, denoting that the word tokens $s_{a_i}$ and $t_{b_i}$ are aligned.
\citet{sabet-etal-2020-simalign} has proposed different algorithms to induce alignment from the same similarity matrix, and the best method varies across language pairs. 
In this work, we consider the relatively conservative (i.e., having higher precision) \textit{argmax} and the higher recall \textit{itermax} algorithm \citep{sabet-etal-2020-simalign}, and denote the alignments by $\mathcal{A}_\textit{argmax}$ and $\mathcal{A}_\textit{itermax}$ respectively. 

We substitute the non-contextualized word similarity feature (Section~\ref{sec:weakly-sup-lexind}) with contextualized word similarity where the corresponding word embedding is computed by averaging the final-layer contextualized subword embeddings of \criss{}. The cosine similarities and dot-products of these embeddings are included as features.

Instead of the binary classification in Section~\ref{sec:weakly-sup-lexind}, we do ternary classification for word alignments. For a pair of word tokens $\langle s_i, t_j \rangle$, the gold label $y_{\langle s_i, t_j\rangle}$ is defined as 
\newcommand\1{\mathbbm{1}}
\begin{align*}
\1 [\langle i,j \rangle \in \mathcal{A}_\textit{argmax} ] +\1 [\langle i,j \rangle \in \mathcal{A}_\textit{itermax} ].
\end{align*}
Intuitively, the labels $0$ and $2$ represents confident alignment or non-alignment by both methods, while the label $1$ models the potential alignment. 

The MLP takes the features $\mathbf{x}_{\langle s_i,t_j \rangle} \in \mathbb{R}^7$ of the word token pair, and compute the probability of each label $y$ by 
\begin{align*}
    \bm{\hat{h}} &= \mathrm{ReLU}\left(\mathbf{W}_1\mathbf{x}_{\langle s_i,t_j\rangle} + \mathbf{b}_1\right) \\
    \bm{g} &= \mathbf{W}_2 \cdot \bm{\hat{h}} + \mathbf{b}_2  \\
    P_\Phi(y \mid  s_i, t_j, \mathcal{S}, \mathcal{T}) &= \frac{\exp\left(g_y\right)}{\sum_{y'}\exp\left(g_{y'}\right)}, 
\end{align*}
where $\Phi = \{\mathbf{W}_1 \mathbf{W}_2, \mathbf{b}_1, \mathbf{b}_2\}$. 
On the training stage, we maximize the log-likelihood of ground-truth labels: 
\begin{align*}
    \Phi^* & = \arg\max_{\Phi} \\
    & \sum_{\langle\mathcal{S}, \mathcal{T}\rangle \in \mathcal{B}}\sum_{s_i \in \mathcal{S}}\sum_{t_j \in \mathcal{T}} \log P_\Phi(y_{\langle s_i,t_j \rangle} \mid s_i,t_j, \mathcal{S}, \mathcal{T}).
\end{align*}
On the inference stage, we keep all word token pairs $\langle s_i, t_j\rangle$ that have 
\begin{align*}
    \mathbb{E}_P[y] := \sum_y y\cdot P(y \mid s_i,t_j,\mathcal{S},\mathcal{T}) > 1
\end{align*}
as the prediction.

\section{Experimental Setup and Baselines}
Throughout our experiments, we use a two-layer perceptron with the hidden size of 8 for both lexicon induction and word alignment. 
We optimize all of our models using Adam \citep{kingma2015adam} with the initial learning rate $5\times 10^{-4}$. 
For our bitext construction methods, we retrieve the best matching sentence or translate the sentences in the source language Wikipedia; for baseline models, we use their default settings.
\begin{table*}[t]
    \centering \small 
    \begin{tabular}{c|ccccccc|ccc}
    \toprule
        \bf Language & \multicolumn{7}{c|}{\emph{Weakly-Supervised}} & \multicolumn{3}{c}{\emph{Unsupervised}} \\
         \bf Pair & \textsc{bucc} & \textsc{VecMap} & \textsc{wm} & \sc gen & \sc gen-n & \sc rtv & \sc gen-rtv & \textsc{VecMap} & \sc gen & \sc rtv \\
    \midrule
de-en   & 61.5 & 37.1 & 71.6 & 70.2 & 67.7 & 73.0 & \bf 74.2 & 22.1 & 62.6 & 66.8 \\
de-fr   & 76.8 & 43.2 & 79.8 & 79.1 & 79.2 & 78.9 & \bf 83.2 & 27.1 & 79.4 & 80.3 \\
en-de   & 54.5 & 33.2 & 62.1 & 62.7 & 59.3 & 64.4 & \bf 66.0 & 33.7 & 51.0 & 56.2 \\
en-es   & 62.6 & 45.3 & 71.8 & 73.7 & 69.6 & \bf 77.0 & 75.3 & 44.1 & 60.2 & 65.6 \\
en-fr   & 65.1 & 45.4 & 74.4 & 73.1 & 69.9 & 73.4 & \bf 76.3 & 44.8 & 61.9 & 66.3 \\
en-ru   & 41.4 & 29.2 & \bf 54.4 & 43.5 & 37.9 & 53.1 & 53.1 & 24.6 & 28.4 & 45.4 \\
en-zh   & 49.5 & 31.0 & 67.7 & 64.3 & 56.8 & \bf 69.9 & 68.3 & 12.8  & 51.5 & 51.7 \\
es-en   & 71.1 & 55.5 & 82.3 & 80.3 & 75.8 & \bf 82.8 & 82.6 & 52.4 & 71.4 & 76.4 \\
fr-de   & 71.0 & 46.2 & \bf 82.1 & 80.0 & 78.7 & 80.9 & 81.7 & 46.0 & 76.4 & 77.3 \\
fr-en   & 53.7 & 51.5 & 80.3 & 79.7 & 76.1 & 80.0 & \bf 83.2 & 50.4 & 72.7 & 75.9 \\
ru-en   & 57.1 & 44.8 & 72.7 & 61.1 & 59.2 & 72.7 & \bf 72.9 & 42.1 & 51.8 & 68.0 \\
zh-en   & 36.9 & 36.1 & \bf 64.1 & 52.6 & 50.6 & 62.5 & 62.5 & 34.4 & 34.3 & 48.1 \\
\midrule 
average & 58.4 & 41.5 & 72.0 & 68.4 & 65.1 & 72.4 & \bf 73.3 & 36.2 & 58.5 & 64.8 \\
    \bottomrule
    \end{tabular}
    \caption{$F_1$ scores ($\times 100$) on the BUCC 2020 test set \citep{rapp-etal-2020-overview}. The best number in each row is {\bf bolded}. }
    \label{tab:lexind-f1}
\end{table*}

For evaluation, we use the BUCC 2020 BLI shared task dataset \citep{rapp-etal-2020-overview} and metric ($F_1$).
Like most recent work, this evaluation is based on MUSE \citep{conneau2017word}.\footnote{\url{https://github.com/facebookresearch/MUSE}}
We primarily report the BUCC evaluation because it considers recall in addition to precision.
However, because most recent work only evaluates on precision, we include those evaluations in Appendix~\ref{sec:p1eval}. 

We compare the following baselines:

\paragraph{\textsc{BUCC}.} Best results from the BUCC 2020 \citep{rapp-etal-2020-overview} for each language pairs, we take the maximum $F_1$ score between the best closed-track results \citep{severini-etal-2020-lmu,laville-etal-2020-taln} and open-track ones \citep{severini-etal-2020-lmu}. Our method would be considered open track since the pretrained models used a much larger data set (Common Crawl 25) than the BUCC 2020 closed-track (Wikipedia or Wacky; \citeauthor{baroni2009wacky}, \citeyear{baroni2009wacky}).
%
\paragraph{\textsc{VecMap}.}  Popular and robust method for aligning monolingual word embeddings via a linear projection and extracting lexicons.
Here, we use the standard implementation%
\footnote{\url{https://github.com/artetxem/VecMap}}
with FastText vectors \cite{bojanowski-etal-2017-enriching}\footnote{\url{https://github.com/facebookresearch/fastText}} trained on the union of Wikipedia and Common Crawl corpus for each language.%
\footnote{\url{https://github.com/facebookresearch/fastText/blob/master/docs/crawl-vectors.md}; that is, our \textsc{VecMap} baselines have the same data availability with our main results.}
We include both supervised and unsupervised versions.
%
\paragraph{\textsc{wm}.} WikiMatrix \citep{schwenk2019wikimatrix}%
\footnote{\url{https://github.com/facebookresearch/LASER/tree/master/tasks/WikiMatrix}}
is a dataset of mined bitext. The mining method LASER \citep{artetxe-schwenk-2019-massively} is trained on real bitext and then used to mine more bitext from the Wikipedia corpora to get the WikiMatrix dataset.
We test our lexicon induction method with WikiMatrix bitext as the input and compare to our methods that do not use bitext supervision.

\section{BLI Results and Analysis} 


\subsection{Main Results}
We evaluate bidirectional translations from beam search (\textsc{gen}; Section~\ref{sec:unsup-bitext-cons-gen}), bidirectional translations from nucleus sampling \citep[\textsc{gen-n};][]{holtzman2019curious},\footnote{We sample from the smallest word set whose cumulative probability mass exceeds 0.5 for next words.} and retrieval (\textsc{rtv}; Section~\ref{sec:unsup-bitext-cons-rtv}).
In addition, it is natural to concatenate the global statistical features (Section~\ref{sec:weakly-sup-lexind}) from both \textsc{gen} and \textsc{rtv} and we refer to this approach by \textsc{gen-rtv}. 

Our main results are presented in Table~\ref{tab:lexind-f1}.
All of our models (\textsc{gen, gen-n, rtv, gen-rtv}) outperform the previous state of the art (\textsc{bucc}) by a significant margin on all language pairs.
Surprisingly, \textsc{rtv} and \textsc{gen-rtv} even outperform WikiMatrix by average $F_1$ score, indicating that we do not need bitext supervision to obtain high-quality lexicons. 

\subsection{Automatic Analysis}

\paragraph{Bitext quality.}
\label{sec:bitext-quality}
\begin{table*}[t]
\centering \small 
\begin{tabular}{c|cccccc|c}
\toprule
&  \multicolumn{6}{|c|}{\it Bitext Quality: High $\rightarrow$ Low} & \\ 
Lang. & \multicolumn{1}{l}{\sc rtv-1} & \multicolumn{1}{l}{\sc rtv-2} & \multicolumn{1}{l}{\sc rtv-3} & \multicolumn{1}{l}{\sc rtv-4} & \multicolumn{1}{l}{\sc rtv-5} &  Random & \sc rtv-all \\
\midrule
de-en & \bf 73.0  & 67.9  & 65.8  & 64.5  & 63.1  & 37.8 & 70.9 \\
de-fr & 78.9  & 74.2  & 70.8  & 69.5  & 67.3  & 60.6 & \bf 79.4 \\
en-de & \bf 64.4  & 59.7  & 58.1  & 56.6  & 57.2  & 36.5 & 62.5 \\
en-es & \bf 77.0  & 76.5  & 73.7  & 68.4  & 66.1  & 43.3 & 75.3 \\
en-fr & \bf 73.4  & 70.5  & 67.9  & 65.7  & 65.5  & 47.8 & 68.3 \\
en-ru & \bf 53.1  & 48.0  & 44.2  & 40.8  & 41.0  & 15.0 & 51.3 \\
en-zh & \bf 69.9  & 59.6  & 66.1  & 60.1  & 61.3  & 48.2 & 67.6 \\
es-en & \bf 82.8  & 82.4  & 79.6  & 74.2  & 72.3  & 44.4 & 81.1 \\
fr-de & \bf 80.9  & 76.9  & 73.2  & 74.7  & 74.5  & 64.7 & 79.1 \\
fr-en & \bf 80.0  & 79.0  & 74.2  & 72.6  & 71.6  & 50.1 & 79.4 \\
ru-en & \bf 72.7  & 66.8  & 60.5  & 55.8  & 54.0  & 14.7 & 71.0 \\
zh-en & \bf 62.5  & 58.0  & 54.1  & 50.9  & 49.3  & 13.6 & 61.3 \\
\midrule 
avg.  & \bf 72.4  & 68.3  & 65.7  & 62.8  & 61.9  & 39.7 & 70.6 \\
\bottomrule
\end{tabular}
\caption{\label{tab:lexind-rtv-quality} $F_1$ scores $(\times 100)$ on the test set of the BUCC 2020 shared task \citep{rapp-etal-2020-overview}. We use the weakly supervised algorithm  (Section~\ref{sec:weakly-sup-lexind}). The best number in each row is bolded. 
\textsc{rtv-1} is the same as \textsc{rtv} in Table~\ref{tab:lexind-f1}.
}
\end{table*}
Since \textsc{rtv} achieves surprisingly high performance, we are interested in how much the quality of bitext affects the lexicon induction performance. 
We divide all retrieved bitexts with score (Eq.~\ref{eq:margin-score}) larger than 1 equally into five sections with respect to the score, and compare the lexicon induction performance (Table~\ref{tab:lexind-rtv-quality}). 
In the table, \textsc{rtv-1} refers to the bitext of the highest quality and \textsc{rtv-5} refers to the ones of the lowest quality, in terms of the margin score (Eq~\ref{eq:margin-score}).%
\footnote{See Appendix~\ref{sec:qualitytiers} for examples from each tier.} 
We also add a random pseudo bitext baseline (Random), where all the bitext are randomly sampled from each language pair, as well as using all retrieved sentence pairs that have scores larger than 1 (\textsc{rtv-all}). 

In general, the lexicon induction performance of \textsc{rtv} correlates well with the quality of bitext. 
Even using the bitext of the lowest quality (\textsc{rtv-5}), it is still able to induce reasonably good bilingual lexicon, outperforming the best numbers reported by BUCC 2020 participants (Table~\ref{tab:lexind-f1}) on average. 
However, \textsc{rtv} achieves poor performance with random bitext (Table~\ref{tab:lexind-rtv-quality}), indicating that it is only robust to a reasonable level of noise. While this is a lower-bound on bitext quality, even random bitext does not lead to 0 $F_1$ since the model may align any co-occurrences of correct word pairs even when they appear in unrelated sentences. 

\paragraph{Word alignment quality.}
\begin{table}[t]
    \centering \small  
    \begin{tabular}{ccc}
    \toprule
         Languages & \simalign & \fastalign \\
    \midrule
        de-en & \bf 73.0 &  69.7 \\
        de-fr & \bf 78.9 &  69.1 \\
        en-de & \bf 64.4 &  61.2 \\
        en-es & \bf 77.0 &  72.8 \\
        en-fr & \bf 73.4 &  68.5 \\
        en-ru & \bf 53.1 &  50.7 \\
        en-zh & \bf 69.9 &  66.0 \\
        es-en & \bf 82.8 &  79.8 \\
        fr-de & \bf 80.9 &  75.8 \\
        fr-en & \bf 80.0 &  77.3 \\
        ru-en & \bf 72.7 &  70.2 \\
        zh-en & \bf 62.5 &  60.2 \\
        \midrule 
        average  & \bf 72.4  & 68.4 \\
    \bottomrule
    \end{tabular}
    \caption{$F_1$ scores ($\times 100$) on the BUCC 2020 test set. 
    Models are trained with the retrieval--based bitext (\textsc{rtv}), in the weakly-supervised setting (Section~\ref{sec:weakly-sup-lexind}.  
    The best number in each row is bolded. }
    \label{tab:lexind-aligner-quality}
\end{table}

We compare the lexicon induction performance using the same set of constructed bitext (\textsc{rtv}) and different word aligners (Table~\ref{tab:lexind-aligner-quality}). 
According to \citet{sabet-etal-2020-simalign}, \simalign\ outperforms \fastalign\ in terms of word alignment.
We observe that such a trend translates to resulting lexicon induction performance well: a significantly better word aligner can usually lead to a better induced lexicon.

\paragraph{Bitext quantity.}
\begin{figure}[t]
    \centering
    \includegraphics[width=0.3\textwidth]{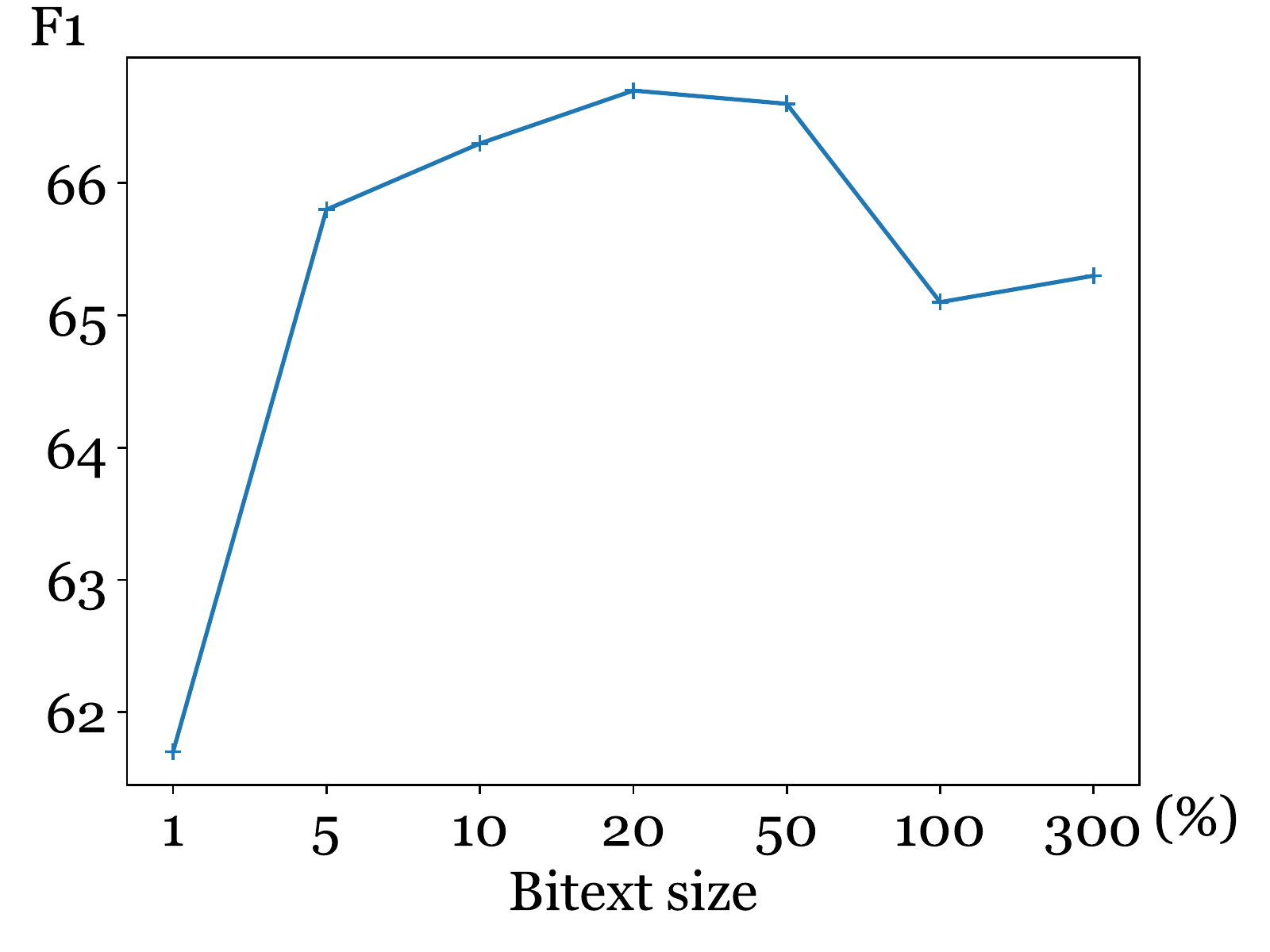}
    \caption{$F_1$ scores ($\times 100$) on the BUCC 2020 test set, produced by our weakly-supervised framework using different amount of bitext generated by \criss\ with nucleus sampling. 100\% is the same as \textsc{gen-n} in Table~\ref{tab:lexind-f1}. For less than 100\%, we uniformly sample the corresponding amount of bitext; for greater, we generate multiple translations for each source sentence. }
    \label{fig:bitext-quantity}
\end{figure}
We investigate how the BLI performance changes when the quantity of bitext changes (Figure~\ref{fig:bitext-quantity}). 
We use \criss{} with nucleus sampling (\textsc{gen-n}) to create different amount of bitext of the same quality. 
We find that with only 1\% of the bitext (160K sentence pairs on average) used by \textsc{gen-n}, our weakly-supervised framework outperforms the previous state of the art (\textsc{bucc}; Table~\ref{tab:lexind-f1}). 
The model reaches its best performance using 20\% of the bitext (3.2M sentence pairs on average) and then drops slightly with even more bitext.
This is likely because more bitext introduces more candidates word pairs.

\paragraph{Dependence on word frequency of \textsc{gen} vs. \textsc{rtv}.}
\begin{figure}[t]
    \centering
    \begin{subfigure}[b]{0.23\textwidth}
        \includegraphics[width=\textwidth]{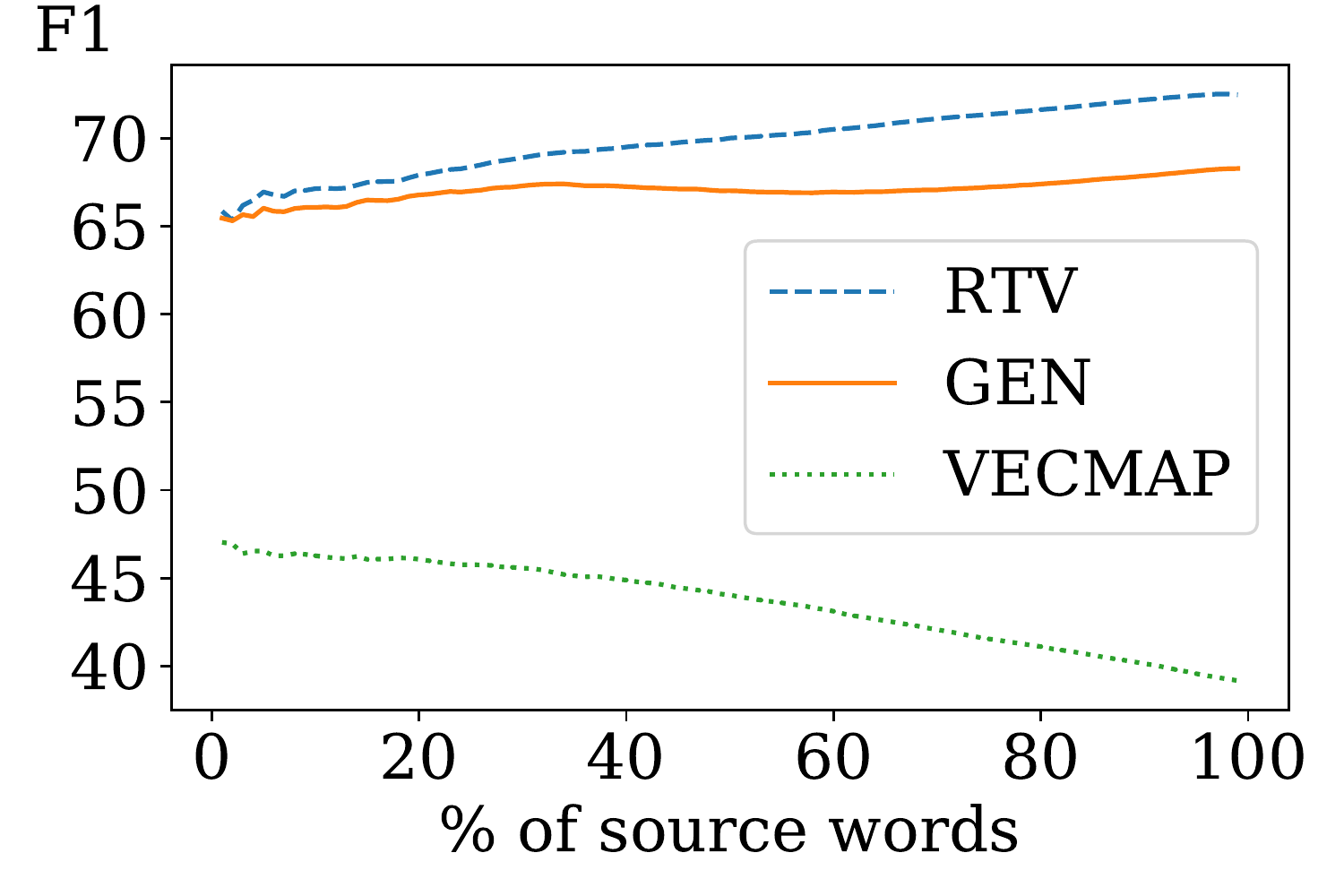}
        \caption{\label{fig:src}}
    \end{subfigure}
    \begin{subfigure}[b]{0.23\textwidth}
        \includegraphics[width=\textwidth]{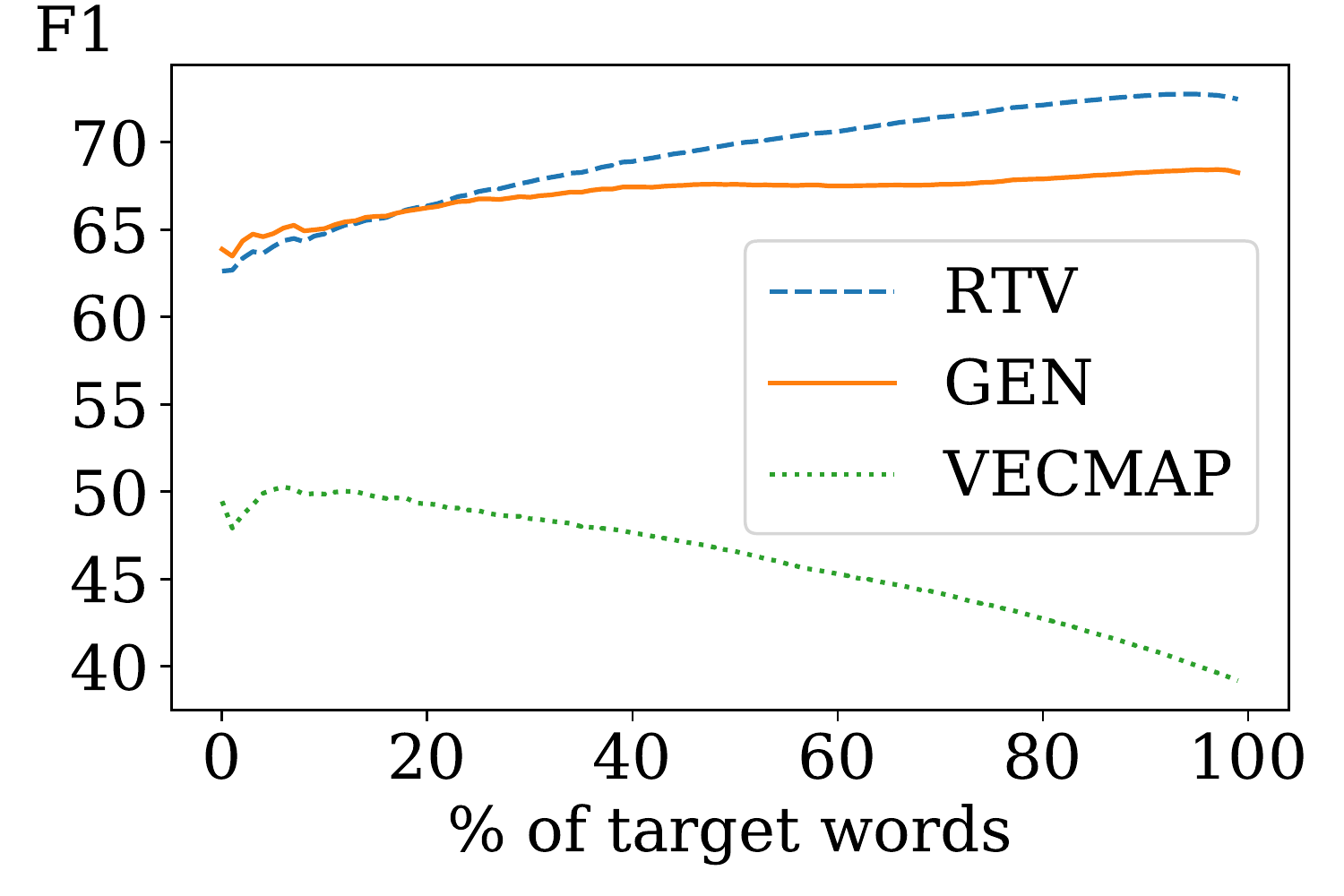}
        \caption{\label{fig:tgt}}
    \end{subfigure}
    \caption{Average $F_1$ scores ($\times 100$) with our weakly-supervised framework across the 12 language pairs (Table~\ref{tab:lexind-f1}) on the filtered BUCC 2020 test set. 
    Results on entries with (a) the k\% most frequent source words, and (b) the k\% most frequent target words. 
    \label{fig:gen-rtv-src-freq}
    }
\end{figure}


We observe that retrieval-based bitext construction (\textsc{rtv}) works significantly better than generation-based ones (\textsc{gen} and \textsc{gen-n}), in terms of lexicon induction performance (Table~\ref{tab:lexind-f1}). 
To further investigate the source of such difference, we compare the performance of the \textsc{rtv} and \textsc{gen} as a function of source word frequency or target word frequency, where the word frequency are computed from the lower-cased Wikipedia corpus. In Figure~\ref{fig:gen-rtv-src-freq}, we plot the $F_1$ of \textsc{rtv} and \textsc{gen} when the most frequent $k\%$ of words are considered.
When all words are considered \textsc{rtv} outperform \textsc{gen} for 11 of 12 language pairs except de-fr.
In 6 of 12 language pairs, 
\textsc{gen} does better than \textsc{rtv} for high frequency source words. As more lower frequency words are included,
\textsc{gen} eventually does worse than \textsc{rtv}. This helps explain why the combined model \textsc{gen-rtv} is even better since \textsc{gen} can have an edge in high frequency words over \textsc{rtv}.
The trend that $F_1(\textsc{rtv}) - F_1(\textsc{gen})$ increases as more lower frequency words are included seems true for all language pairs (Appendix~\ref{sec:langtrends}).

On average and for the majority of language pairs, both methods do better on low-frequency source words than high-frequency ones (Figure~\ref{fig:src}), which is consistent with the findings by BUCC 2020 participants \cite{rapp-etal-2020-overview}.

\paragraph{\textsc{VecMap}.} While BLI through bitext construction and word alignment clearly achieves superior performance than that through vector rotation (Table~\ref{tab:lexind-f1}), we further show that the gap is larger on low-frequency words (Figure~\ref{fig:gen-rtv-src-freq}). 

\subsection{Ground-truth Analysis}
\begin{table}[t]
    \centering \small 
    \begin{tabular}{lc|lc}
    \toprule
    \multicolumn{2}{c|}{\textsc{gen-rtv}} & \multicolumn{2}{c}{\textsc{VecMap}} \\
    \midrule 
倉庫 \hfill depot	&	\cmark	&	申明  \hfill  endorsing	&	\xmark	\\
浪費 \hfill wasting	&	\cmark	&	條件~  \hfill preconditions	&	\bf ?	\\
背面 \hfill reverse	&	\cmark	&	移動  \hfill  moving	&	\cmark	\\
嘴巴 \hfill mouths	&	\cmark	&	天津  \hfill  shanghai	&	\xmark	\\
可笑 \hfill laughable	&	\cmark	&   個案 \hfill cases	&	\cmark	\\
隱藏 \hfill conceal	&	\cmark	&	百合  \hfill  peony	&	\xmark	\\
虔誠 \hfill devout	&	\cmark	&   申報  \hfill  filing	&	\cmark	\\
純淨 \hfill purified	&	\bf ?	&	車廂  \hfill  carriages	&	\cmark	\\
截止 \hfill deadline	&	\cmark	&	海草  \hfill  seaweed	&	\cmark	\\
對外 \hfill foreign	&	\bf ?	&	履歷  \hfill  résumé	&	\cmark	\\
鍾 \hfill clocks	&	\cmark	&	收容所  \hfill  asylums	& \cmark	\\
努力 \hfill effort	&	\cmark	&	開幕  \hfill  soft-opened	&	\xmark	\\
艦 \hfill ships	&	\cmark	&	有形  \hfill  intangible	&	\xmark	\\
州 \hfill states	&	\cmark	&	小刀  \hfill  penknife	&	\cmark	\\
受傷 \hfill wounded	&	\cmark	&	黑山  \hfill  carpathian	&	\cmark	\\
滑動 \hfill sliding	&	\cmark	&	象徵  \hfill  symbolise	&	\cmark	\\
毒理學~ \hfill toxicology	&	\cmark	&	精華  \hfill  fluff-free	&	\xmark	\\
推翻 \hfill overthrown	&	\cmark	&	同謀  \hfill  conspirator	&	\cmark	\\
穿 \hfill wore	&	\cmark	&	籌碼  \hfill  bargaining	&	\xmark	\\
禮貌 \hfill courteous	&	\cmark	&	刮刀  \hfill  rollers	&	\xmark	\\
    \bottomrule 
    \end{tabular}
    \caption{Manually labeled acceptability judgments for random 20 error cases made by \textsc{gen-rtv} (left) and \textsc{VecMap} (right). \cmark\ and \xmark\ denote acceptable and unacceptable translation respectively.  \textbf{?} denotes word pairs that may be acceptable in rare or specific contexts. }
    \label{tab:error-cases}
\end{table}

Following the advice of \citet{kementchedjhieva-etal-2019-lost} that some care is needed due to the incompleteness and biases of the evaluation, we perform manual analysis of selected results.
For Chinese--English translations, we uniformly sample 20 wrong lexicon entries according to the evaluation for both \textsc{gen-rtv} and weakly-supervised \textsc{VecMap}. Our judgments of these samples are shown in Table~\ref{tab:error-cases}.
For \textsc{gen-rtv}, 18/20 of these sampled errors are actually acceptable translations,
whereas for \textsc{VecMap}, only 11/20 are acceptable.
This indicates that the improvement in quality may be partly limited by the incompleteness of the reference lexicon and the ground truth performance of our method might be even better.
The same analysis for English--Chinese is in Appendix~\ref{sec:error-cases-enzh}.

Furthermore, we randomly sample 200 source words from the MUSE zh-en test set, and compare the quality between MUSE translation and those predicted by \textsc{gen-rtv}.
This comparison is MUSE-favored since only MUSE source words are included.
Concretely, we take the union of word pairs, construct the new ground-truth by manual judgments (i.e., removing unacceptable pairs), and evaluate the $F_1$ score against the constructed ground-truth (Table~\ref{tab:manual-judgements-gold}).
The overall gap of 3 $F_1$ means that a higher quality benchmark is necessary to resolve further improvements over \textsc{gen-rtv}.
The word pairs and judgments are included in the supplementary material (Section~\ref{appendix:manual-judgments}).
\begin{table}[t]
    \centering \small 
    \begin{tabular}{crrr}
        \toprule
        Data Source     & Precision & Recall    & $F_1$ \\
        \midrule 
        MUSE            & 93.4      & \bf 78.8      & \bf 85.5  \\
        GEN-RTV         & \bf 96.6      & 71.9      & 82.5  \\
        \bottomrule
    \end{tabular}
    \caption{Comparison of Chinese-English lexicons against manually labeled ground truth. The best number in each column is bolded. }
    \label{tab:manual-judgements-gold}
\end{table}

\section{Word Alignment Results}
\begin{table}[t]
    \centering \small 
    \begin{tabular}{lcccc}
    \toprule 
         \bf Model &  de-en & en-fr & en-hi & ro-en \\
    \midrule 
        \gizapp$^{\dagger}$ & 0.22 & 0.09 & 0.52 & 0.32 \\
        \fastalign$^{\dagger}$ & 0.30 & 0.16 & 0.62 & 0.32\\
        \citet{garg-etal-2019-jointly} & 0.16 & 0.05 &  N/A & 0.23\\
        \citet{zenkel2019adding} & 0.21 & 0.10 & N/A & 0.28 \\ 
    \midrule
        \multicolumn{5}{l}{\simalign\ \citep{sabet-etal-2020-simalign}} \\ 
        \quad \xlmr-\emph{argmax}$^{\dagger}$ & 0.19 & 0.07 & 0.39 & 0.29 \\
        \quad \mbart-\emph{argmax} & 0.20 & 0.09 & 0.45 & 0.29 \\
        \quad \criss-\emph{argmax}$^{*}$ & 0.17 & 0.05 & 0.32 & 0.25 \\
        \quad \criss-\emph{itermax}$^{*}$ & 0.18 & 0.08 & 0.30 & 0.23 \\ 
        MLP (ours)$^{*}$ & \bf 0.15 & \bf 0.04 & \bf 0.28 & \bf 0.22 \\
    \bottomrule
    \end{tabular}
    \caption{Average error rate (AER) for word alignment (lower is better). The best numbers in each column are bolded. Models in the top section require ground-truth bitext, while those in the bottom section do not. $*$: models that involve unsupervised bitext construction. $\dagger$: results copied from \citet{sabet-etal-2020-simalign}. }
    \label{tab:word-alignment}
\end{table}
We evaluate different word alignment methods (Table~\ref{tab:word-alignment}) on existing word alignment datasets,\footnote{\url{http://www-i6.informatik.rwth-aachen.de/goldAlignment}~(de-en);~\url{https://web.eecs.umich.edu/~mihalcea/wpt} (en-fr and ro-en); \url{https://web.eecs.umich.edu/~mihalcea/wpt05} (en-hi)}  following \citet{sabet-etal-2020-simalign}. 
We investigate four language pairs: German--English (de-en), English--French (en-fr), English--Hindi (en-hi) and Romanian--English (ro-en).
We find that the \criss-based \simalign\ already achieves competitive performance with the state-of-the-art method \citep{garg-etal-2019-jointly} which requires real bitext for training. 
By ensembling the \emph{argmax} and \emph{itermax}  \criss-based \simalign\ results (Section~\ref{sec:word-alignment}), we set the new state of the art of word alignment without using any bitext supervision. 

However, by substituting the \criss-based \simalign\ in the BLI pipeline with our aligner, we obtain an average $F_1$ score of 73.0 for \textsc{gen-rtv}, which does not improve over the result of 73.3 achieved by \criss-based \simalign\ (Table~\ref{tab:lexind-f1}), indicating that further effort is required to take the advantage of the improved word aligner. 
\section{Discussion}
We present a direct and effective framework for BLI with unsupervised bitext mining and word alignment, which sets a new state of the art on the task. 
From the perspective of pretrained multilingual models \citep[\emph{inter alia}]{conneau2019unsupervised,liu2020multilingual,tran-etal-2020-cross}, our work shows that they have successfully captured information about word translation that can be extracted using similarity based alignment and refinement.
Although BLI is only about word types, it strongly benefits from contextualized reasoning at the token level.


\section*{Acknowledgment}
We thank Chau Tran for help with pretrained CRISS models, as well as Mikel Artetxe, Kevin Gimpel, Karen Livescu, Jiayuan Mao and anonymous reviewers for their valuable feedback on this work. 

\bibliographystyle{acl_natbib}
\bibliography{acl2021}

\appendix
\clearpage
\appendix
{\Large \bf Appendices}
\section{Language-Specific Analysis}
\label{sec:langtrends}
\begin{figure*}[!ht]
\includegraphics[width=0.23\textwidth]{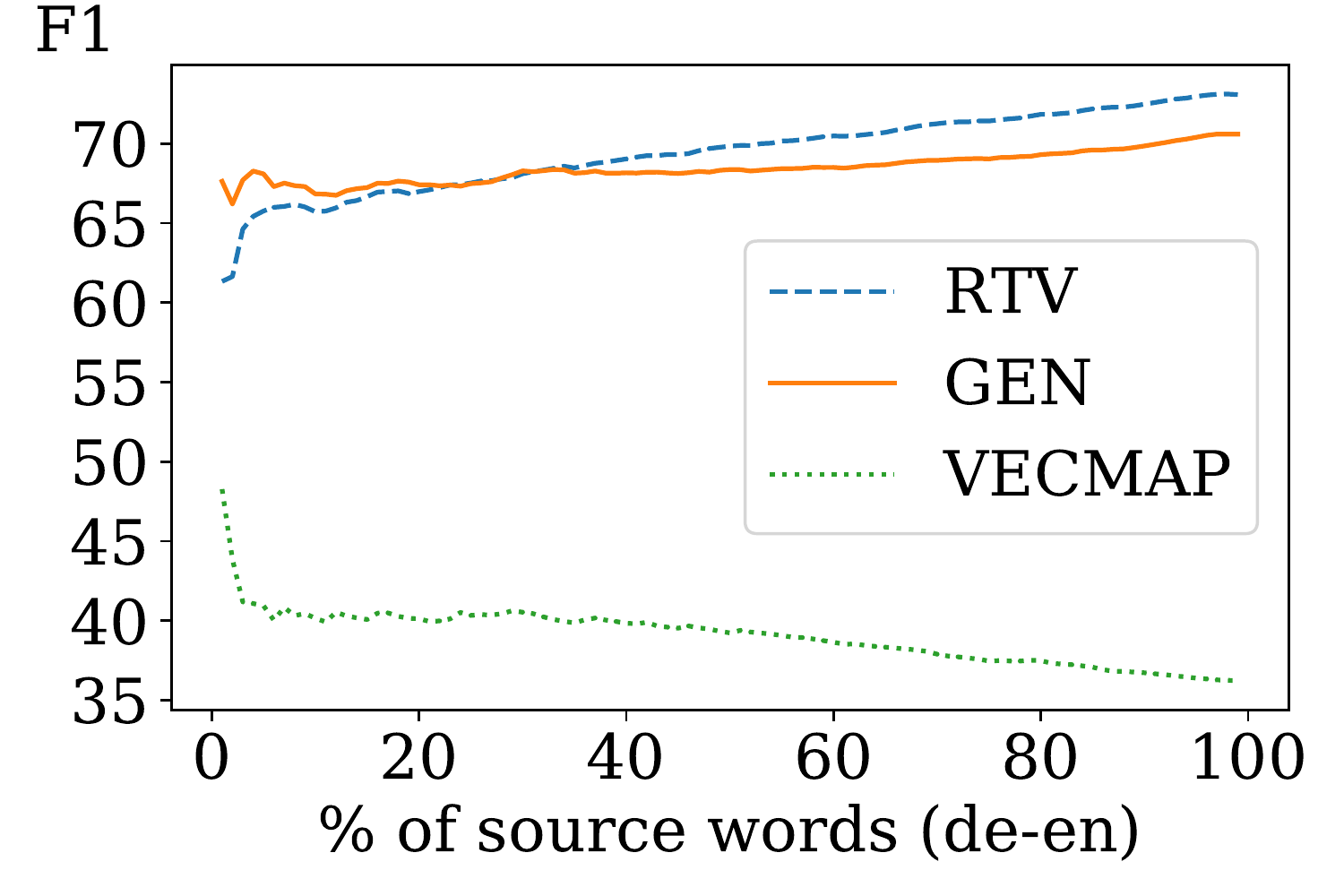}
\includegraphics[width=0.23\textwidth]{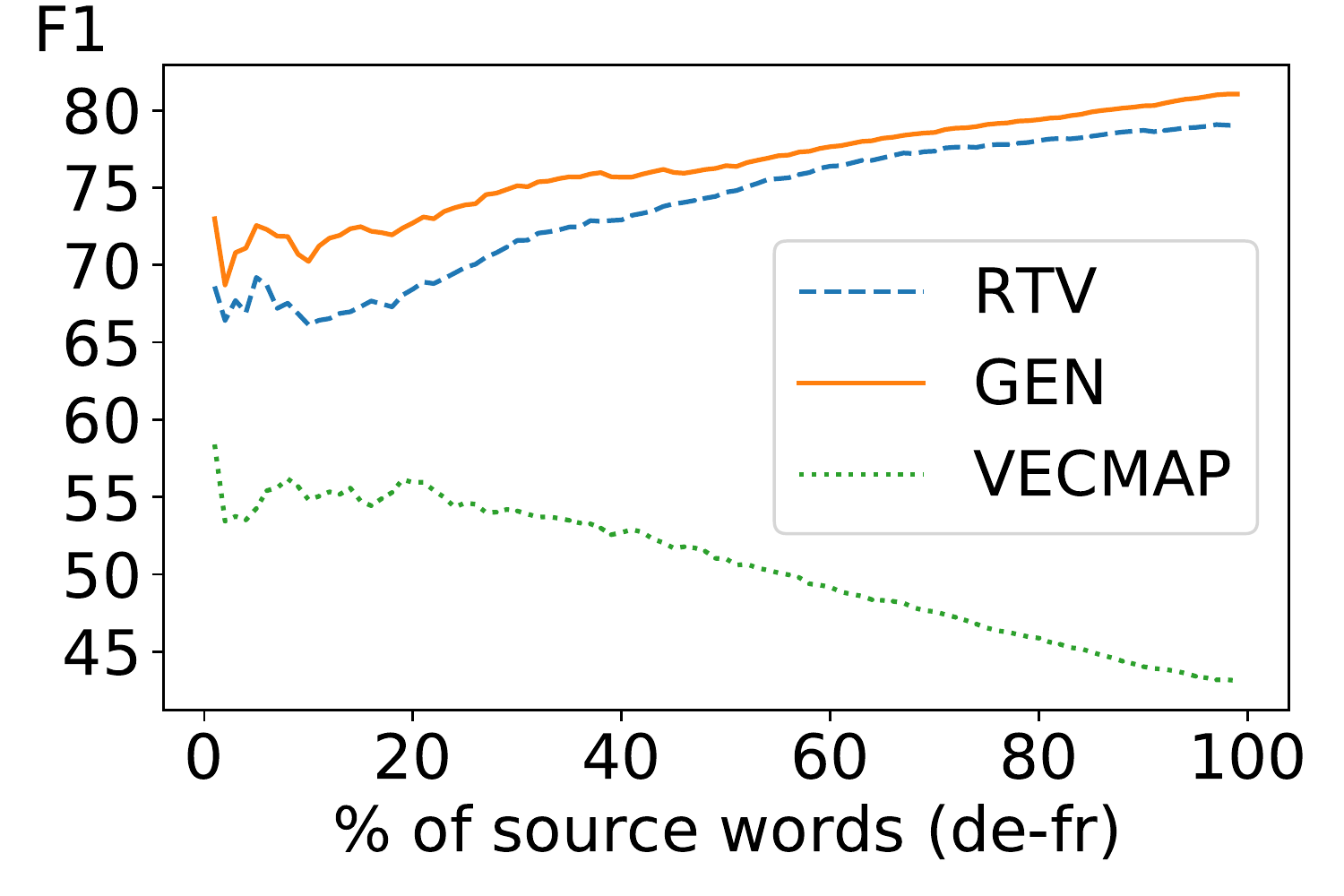}
\includegraphics[width=0.23\textwidth]{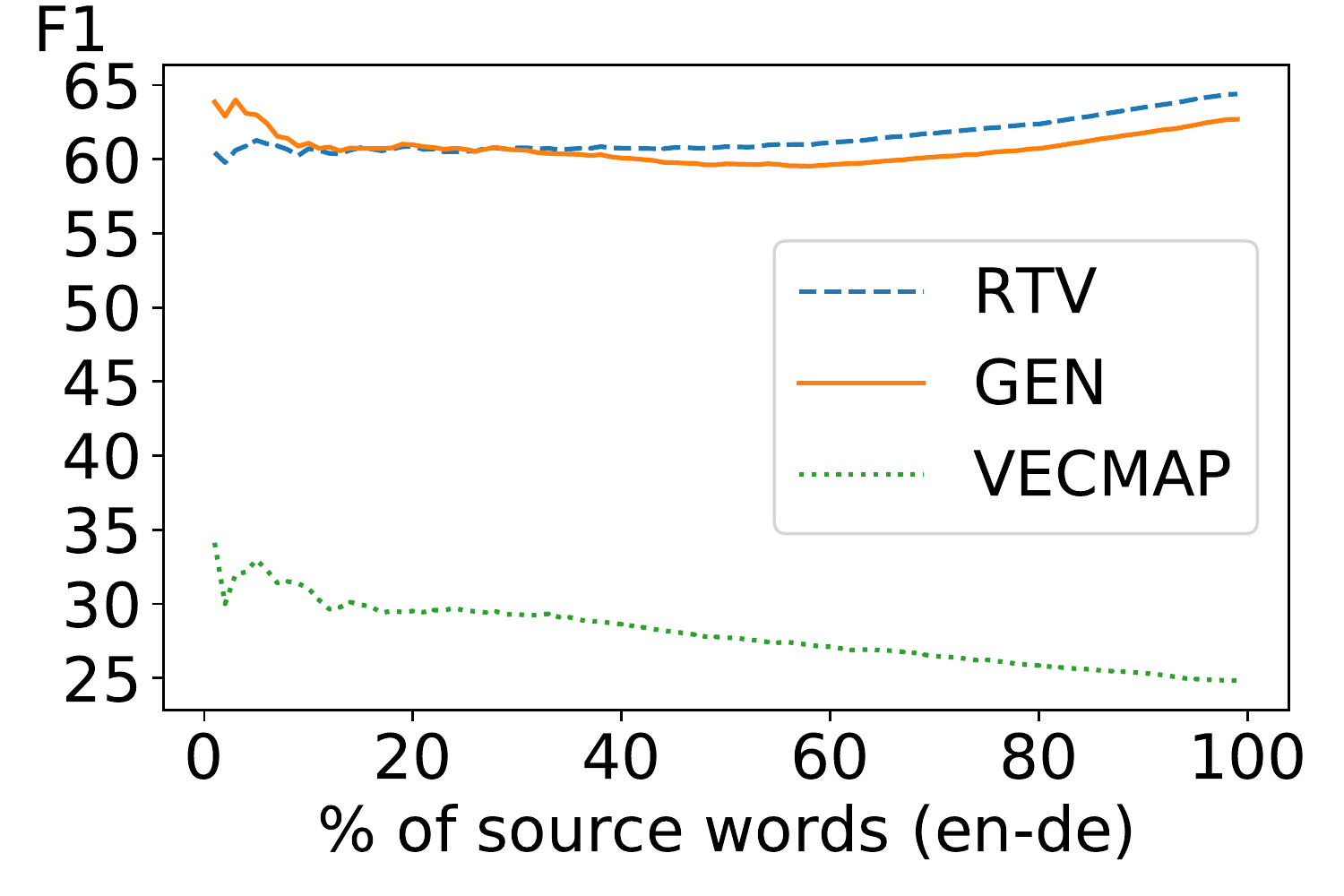}
\includegraphics[width=0.23\textwidth]{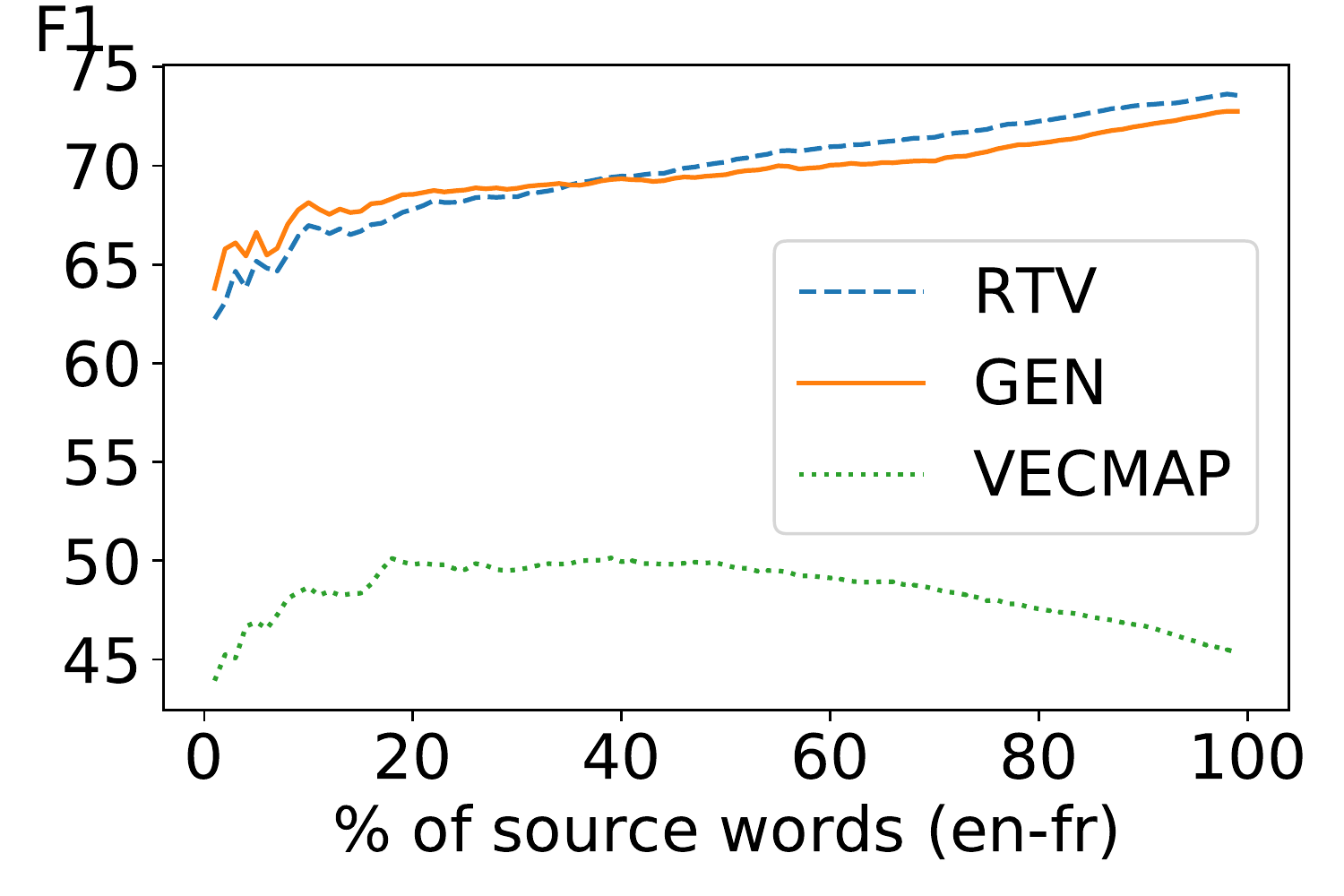} \\
\includegraphics[width=0.23\textwidth]{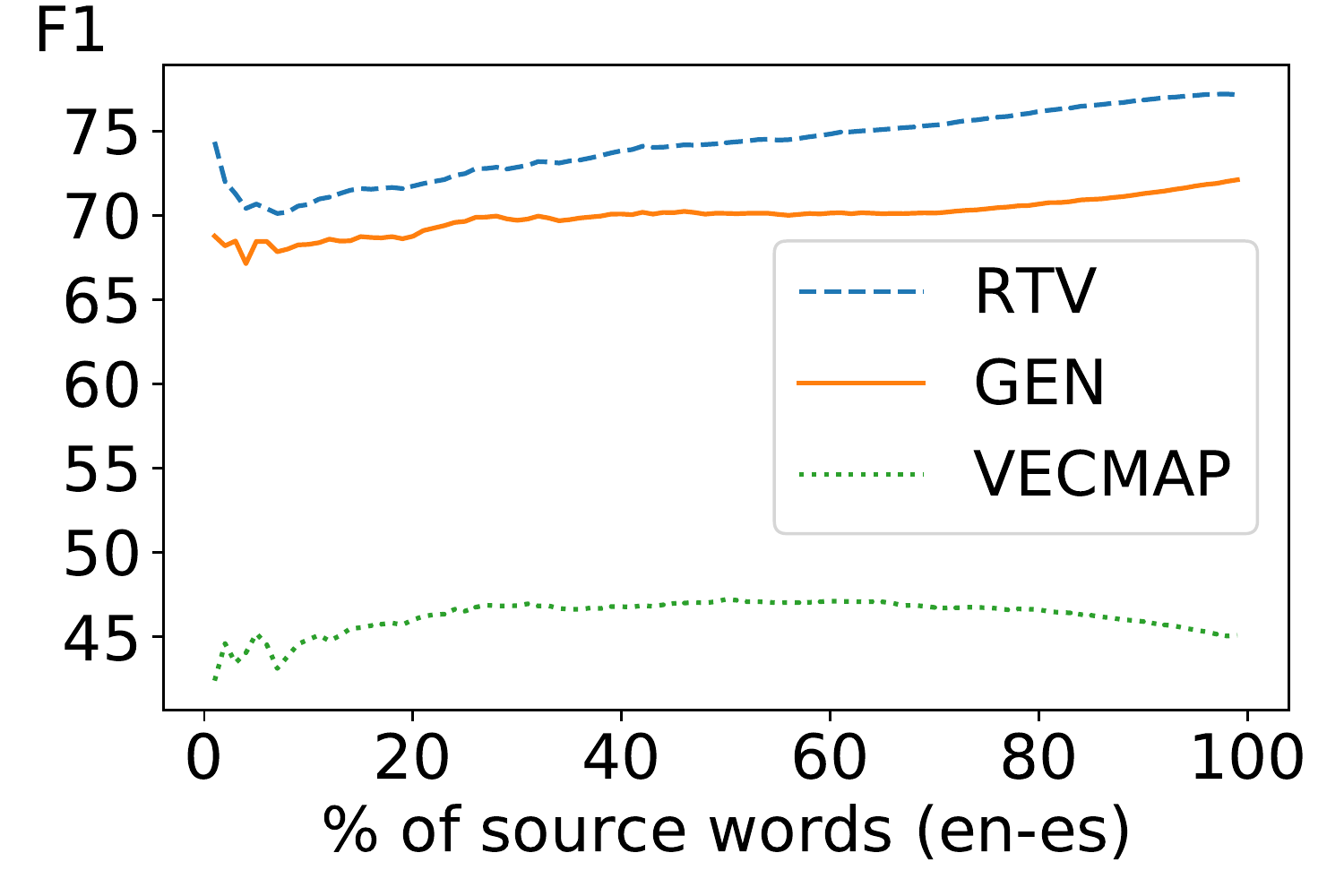}
\includegraphics[width=0.23\textwidth]{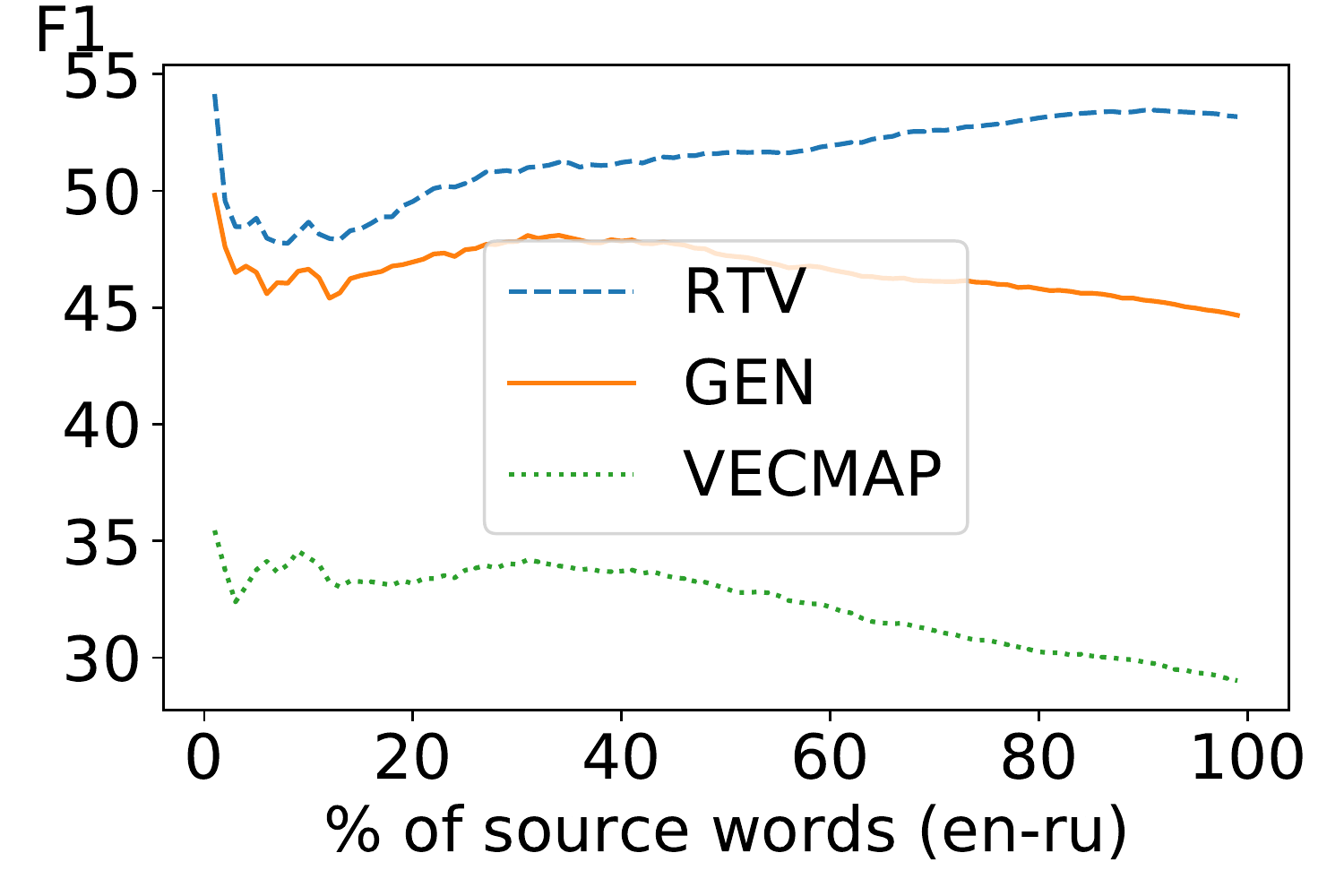}
\includegraphics[width=0.23\textwidth]{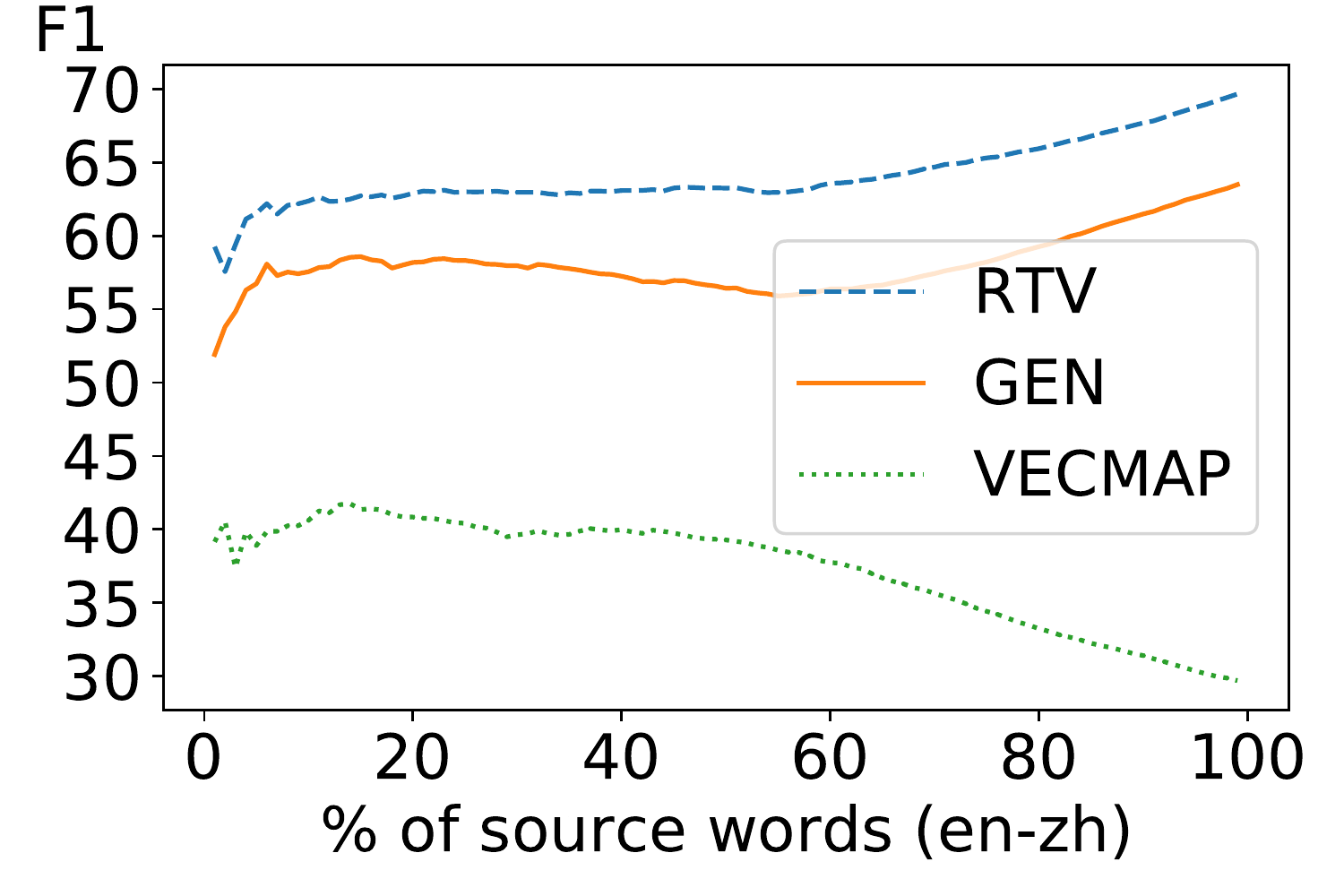}
\includegraphics[width=0.23\textwidth]{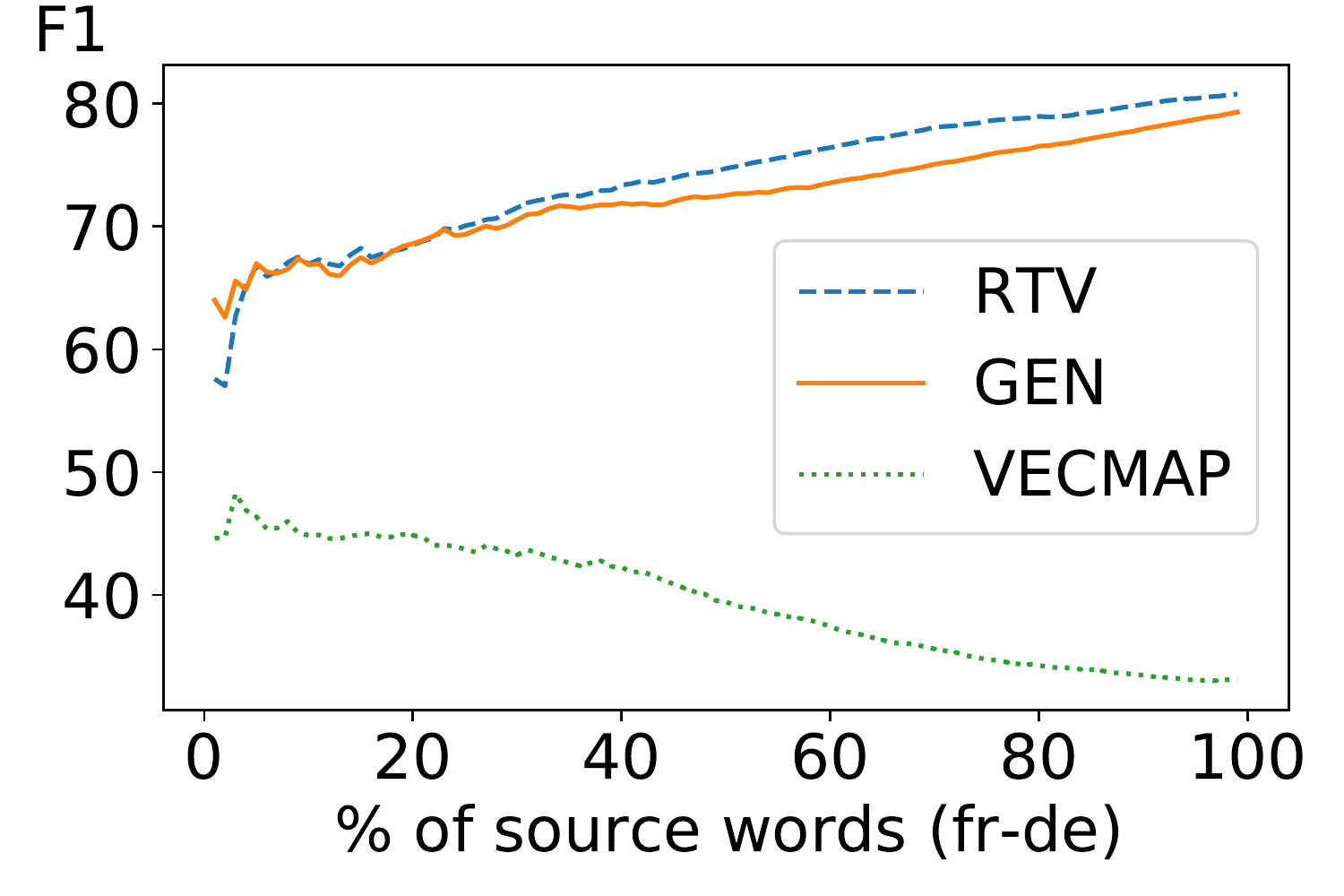} \\
\includegraphics[width=0.23\textwidth]{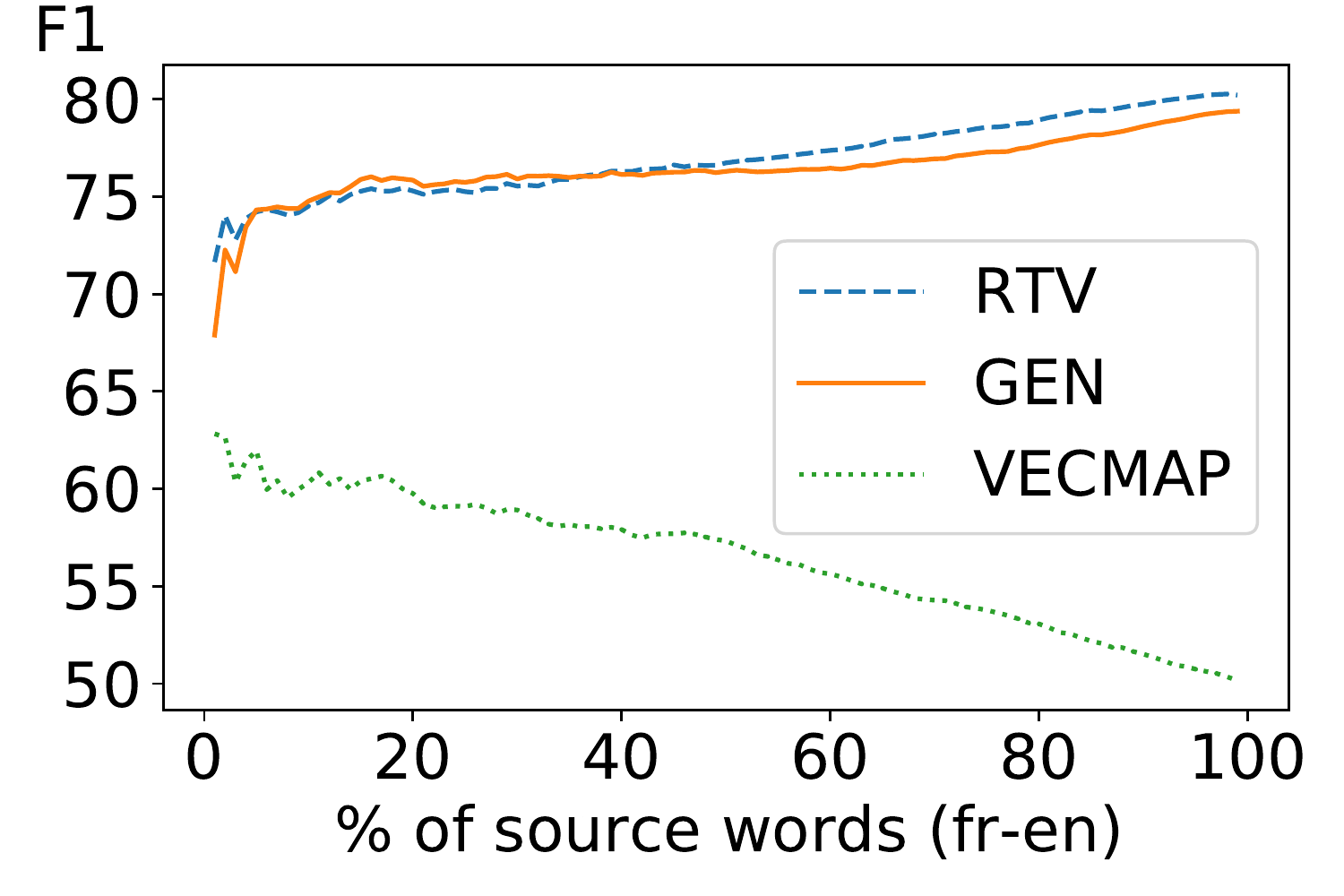}
\includegraphics[width=0.23\textwidth]{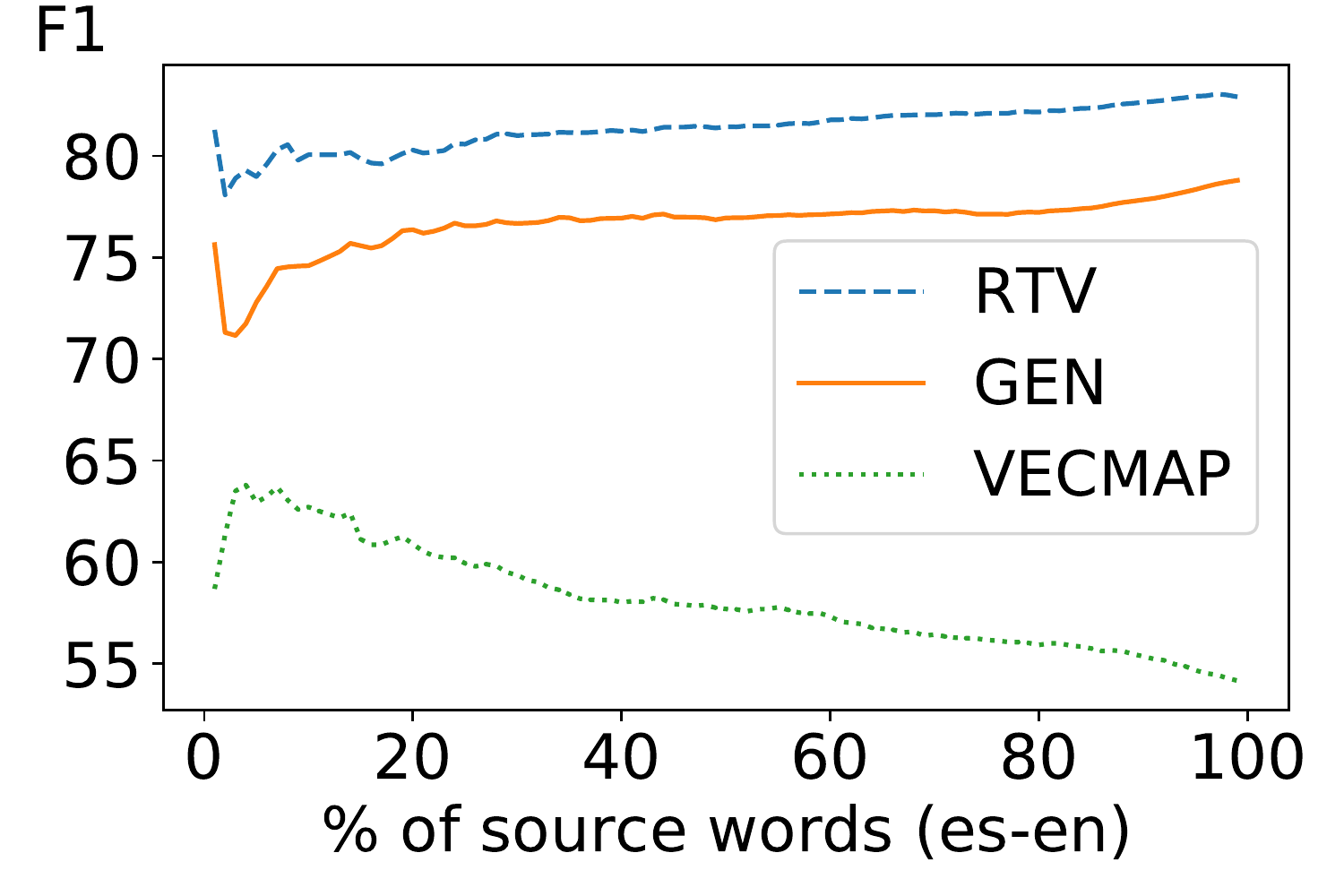}
\includegraphics[width=0.23\textwidth]{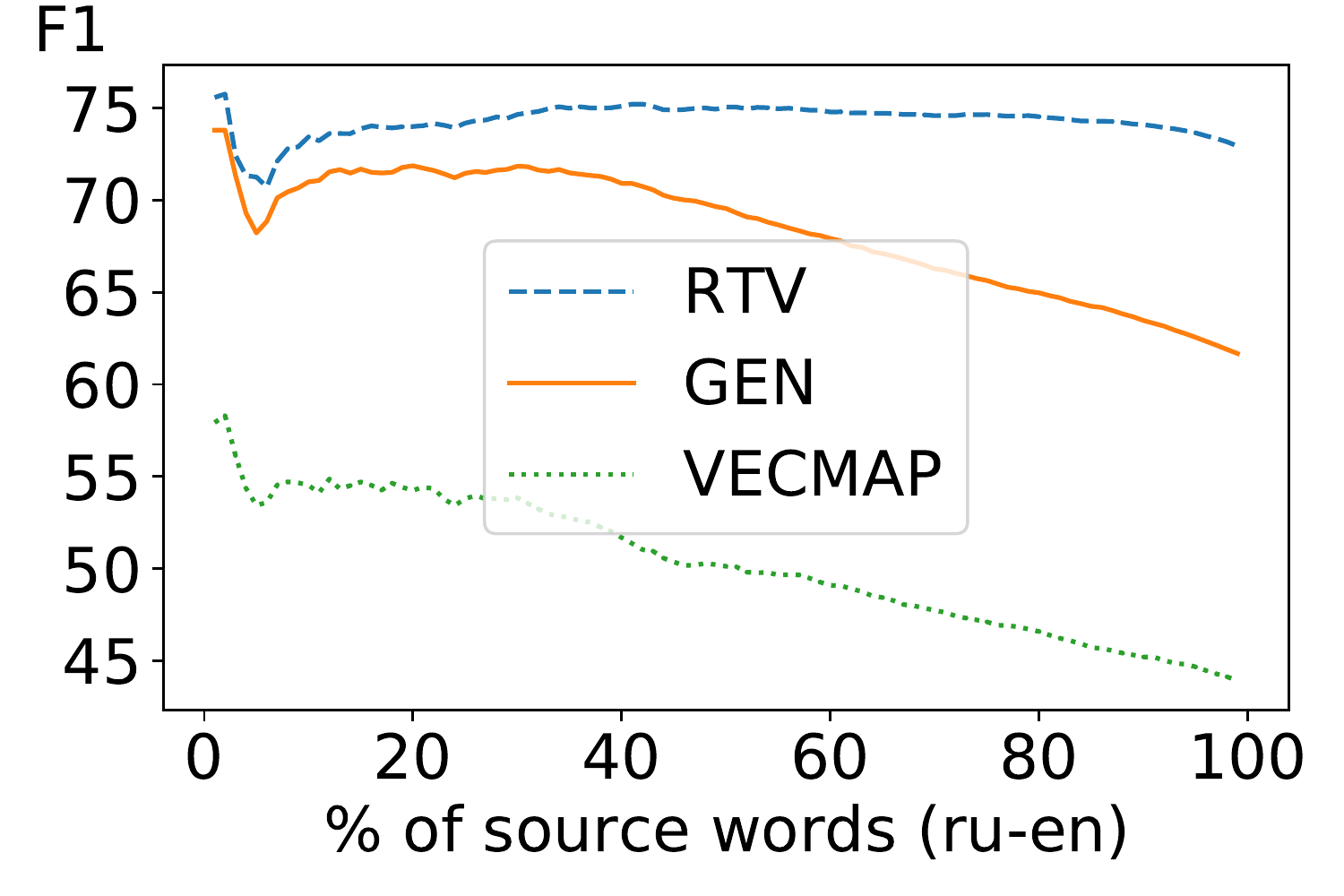}
\includegraphics[width=0.23\textwidth]{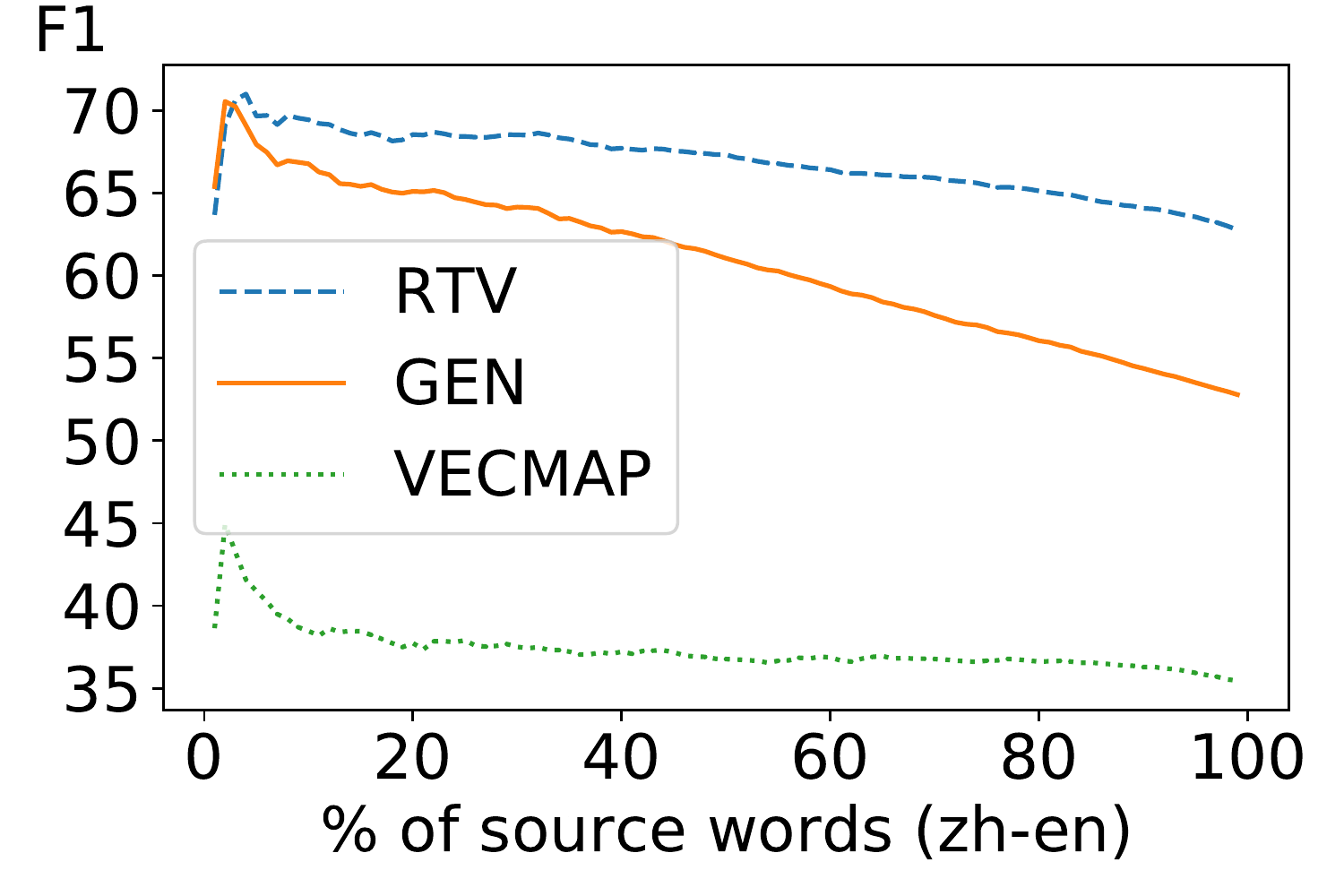} 
\caption{\label{fig:lang-specific-src} $F_1$ scores with respect to portion of source words kept for each investigated language pair, analogous to Figure~\ref{fig:src}.}
\end{figure*}
\begin{figure*}[!bt]
\includegraphics[width=0.23\textwidth]{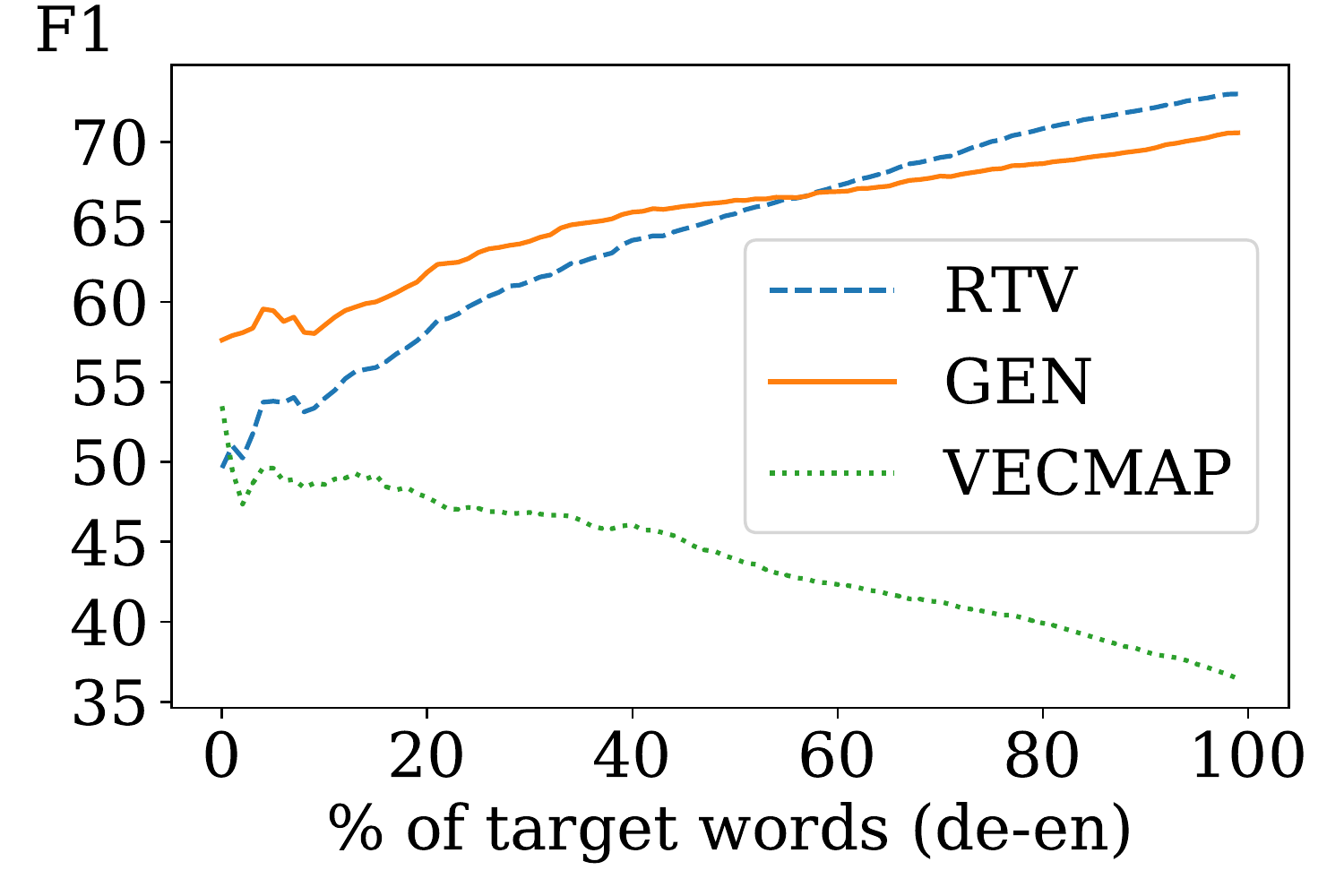}
\includegraphics[width=0.23\textwidth]{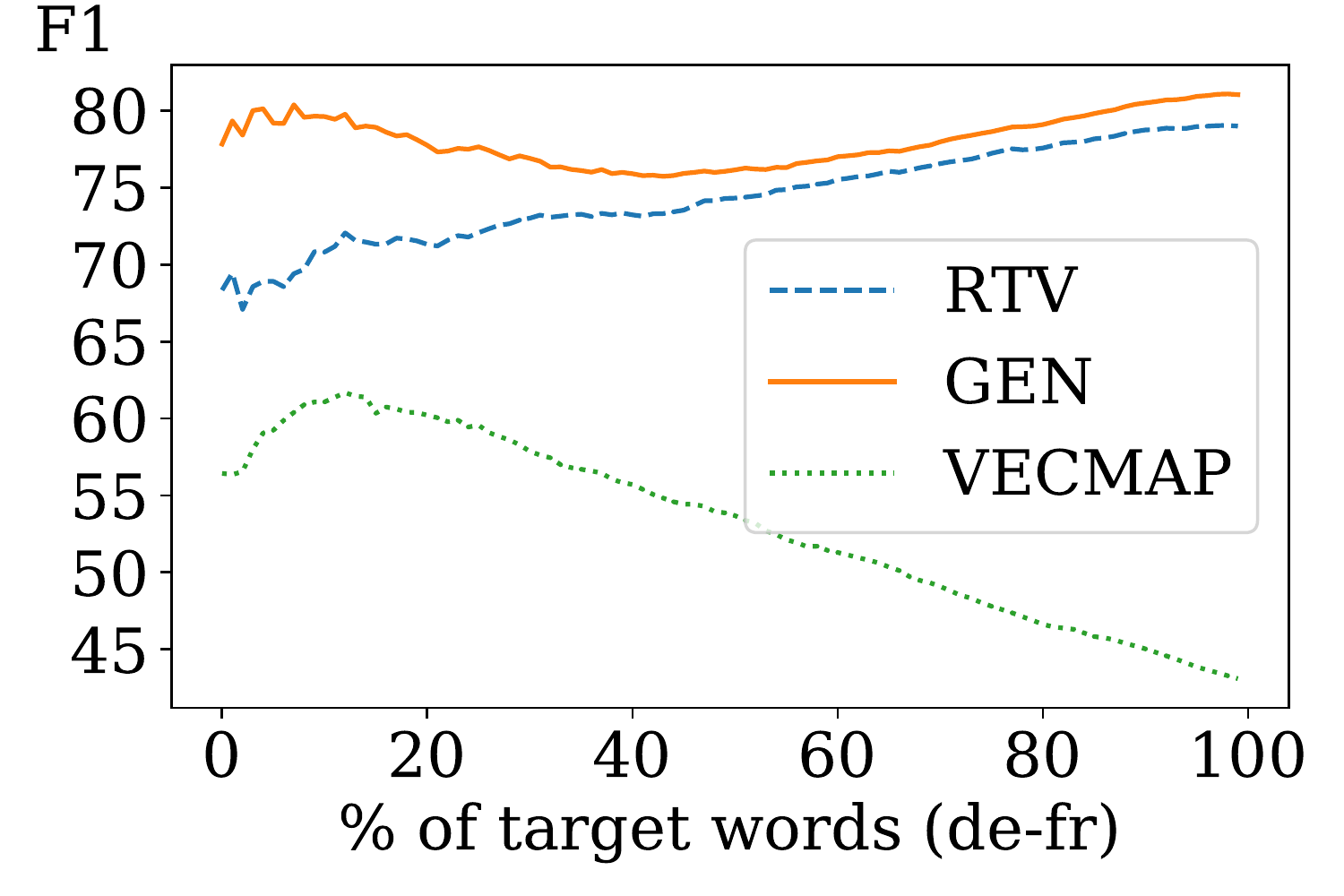}
\includegraphics[width=0.23\textwidth]{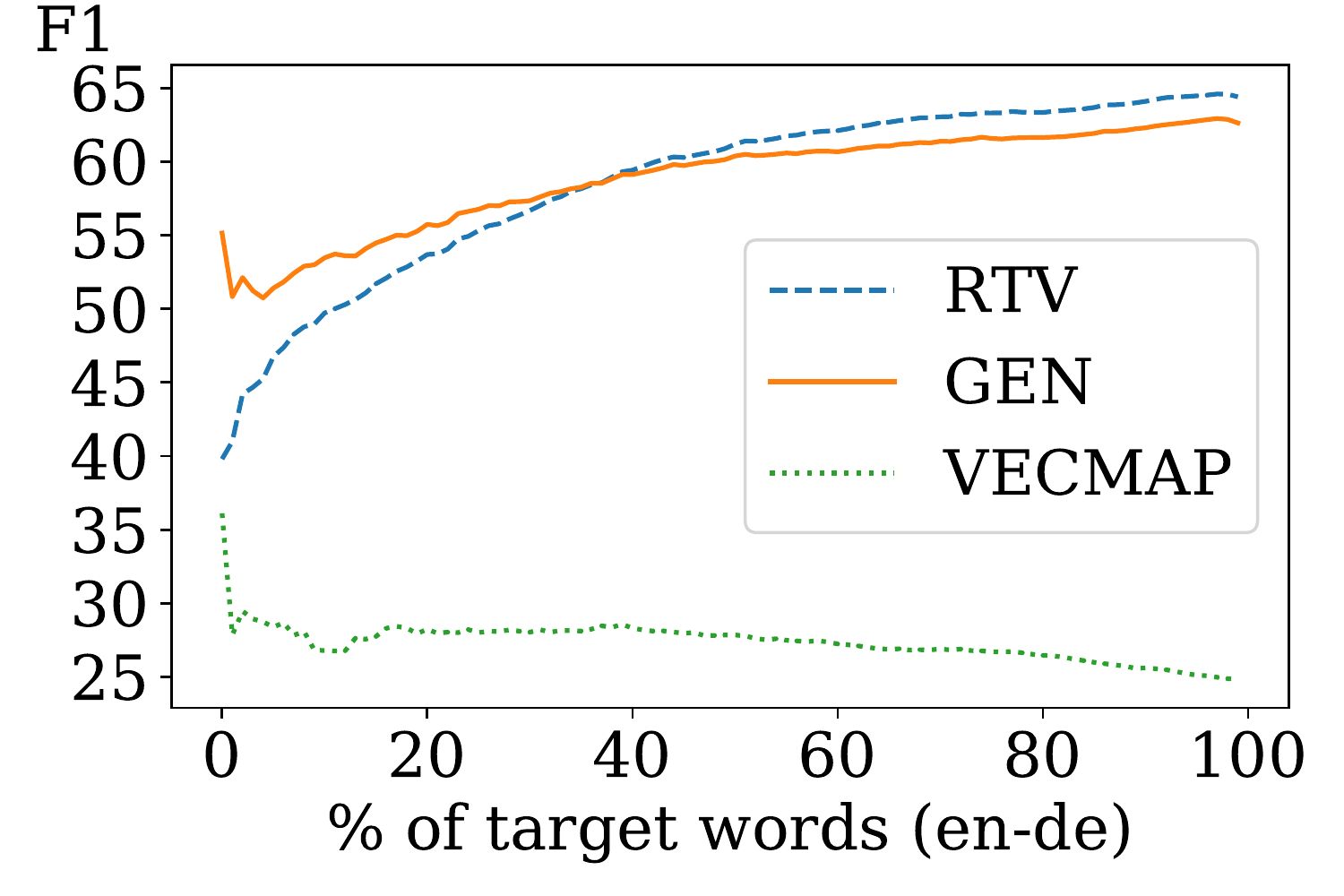}
\includegraphics[width=0.23\textwidth]{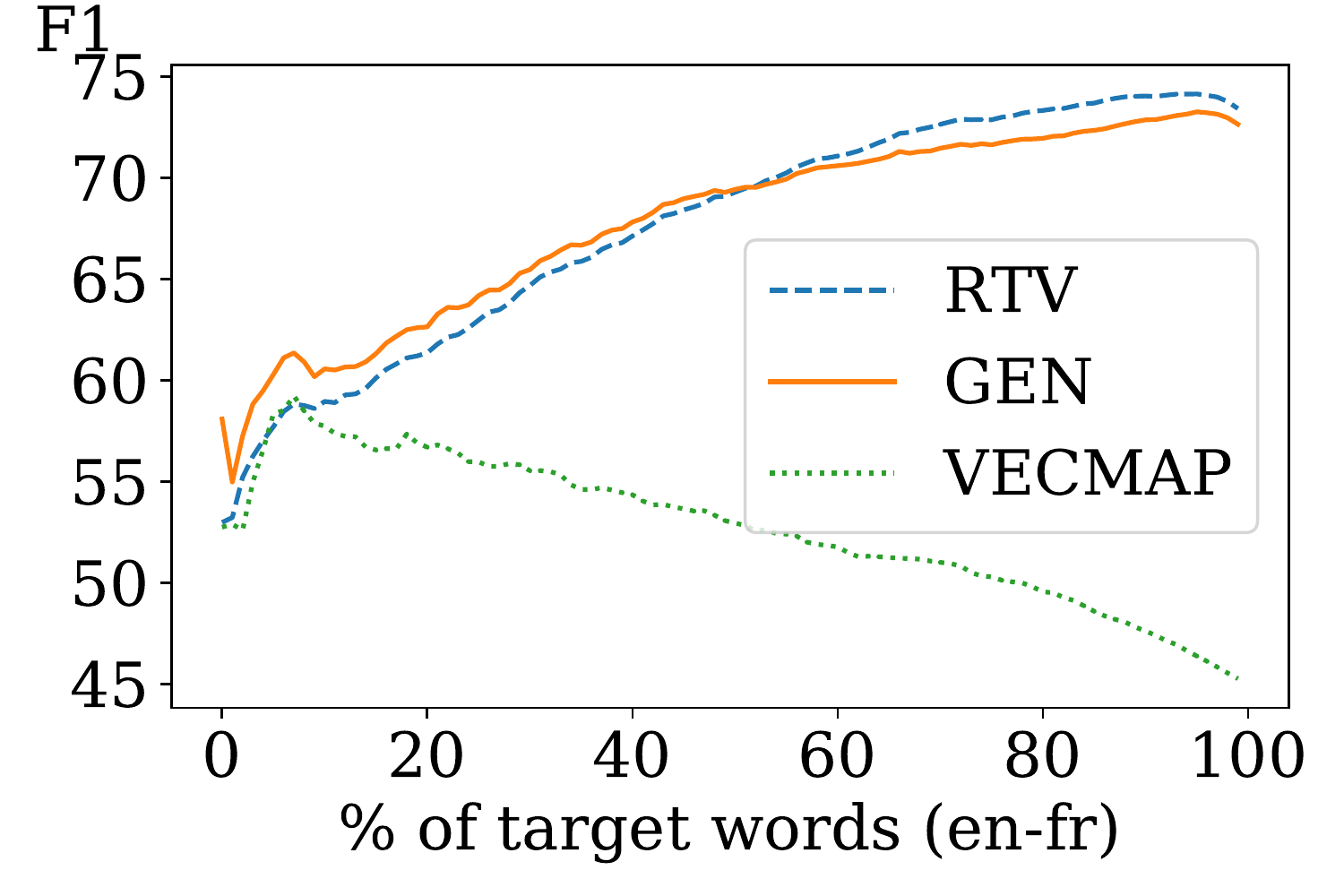} \\
\includegraphics[width=0.23\textwidth]{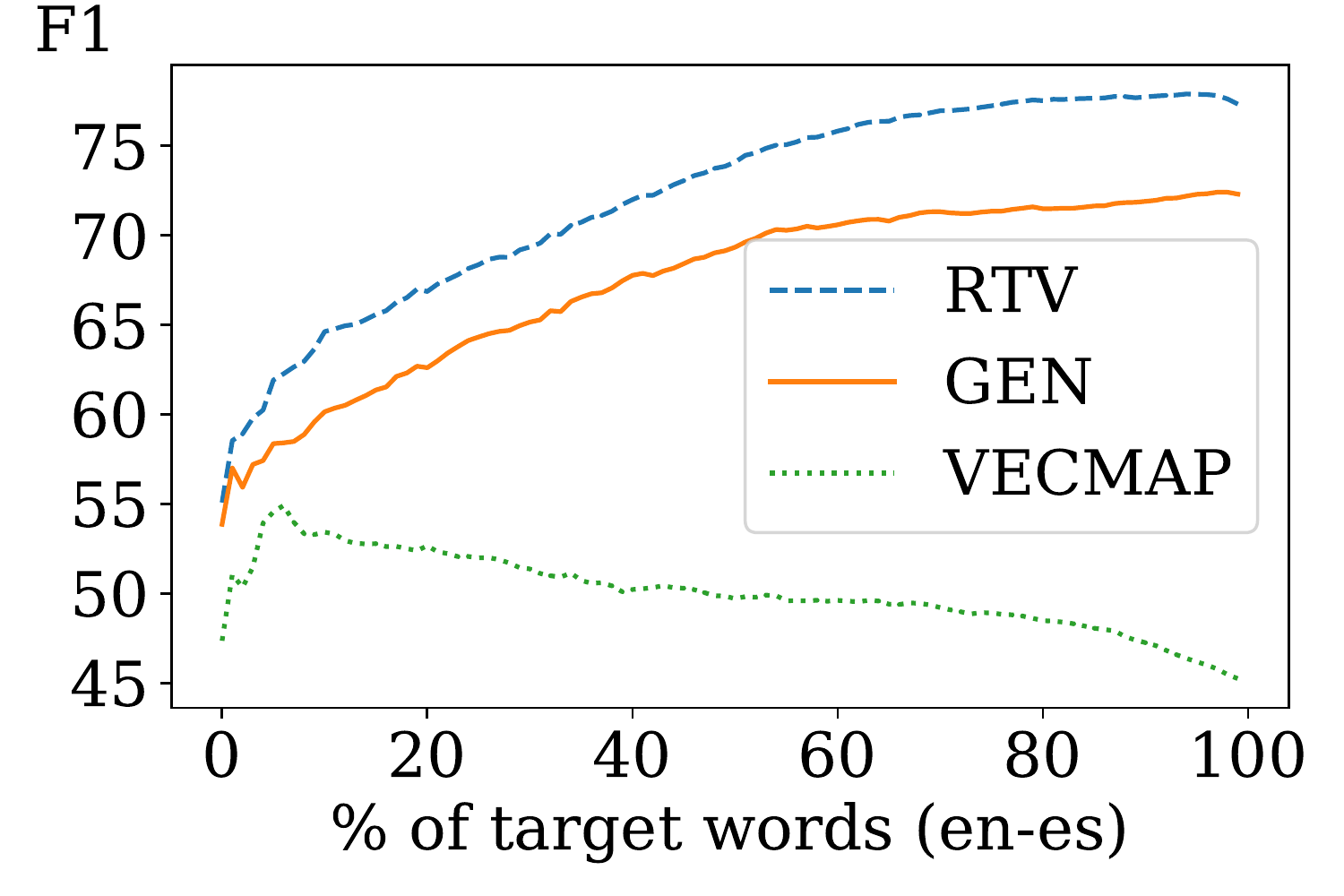}
\includegraphics[width=0.23\textwidth]{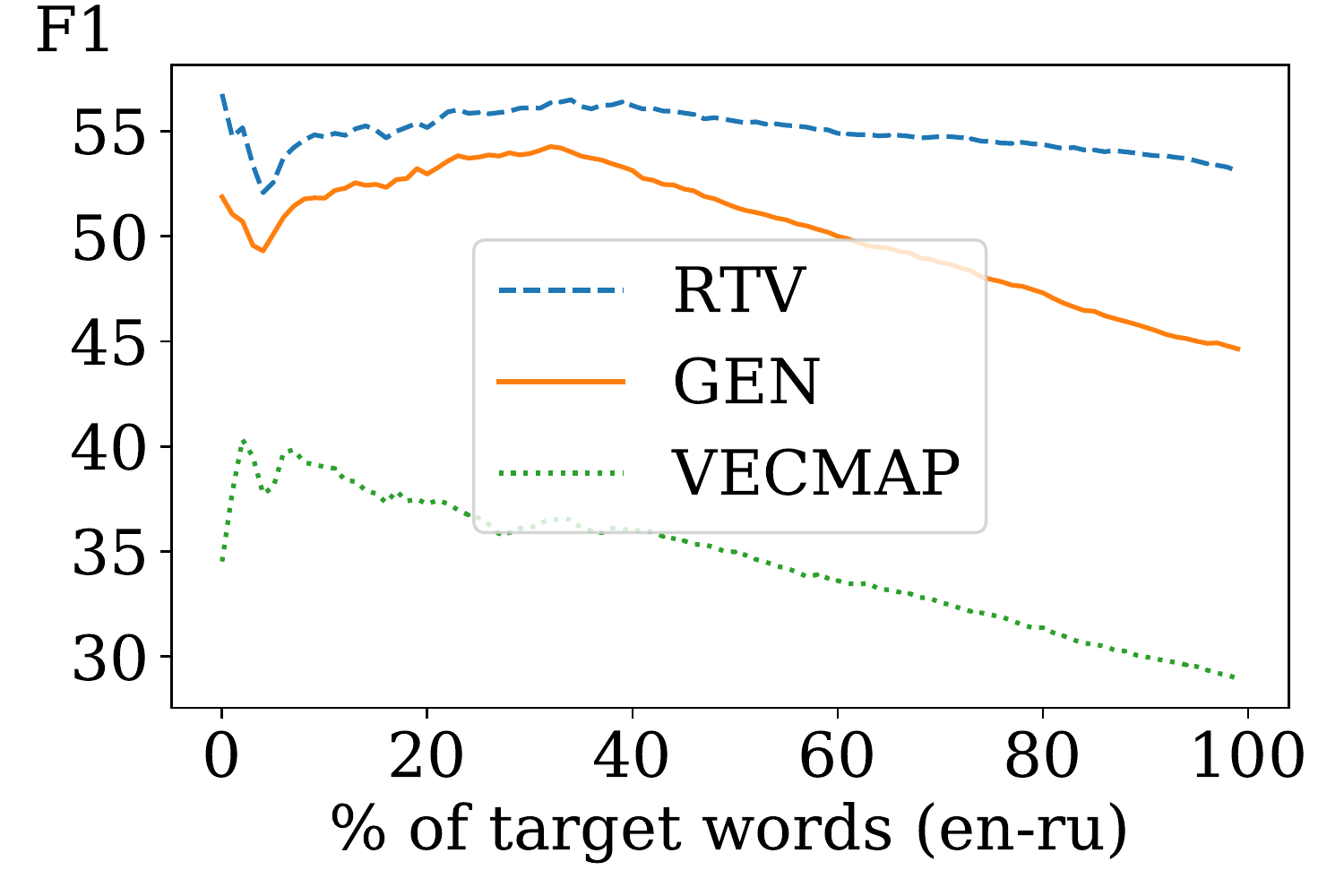}
\includegraphics[width=0.23\textwidth]{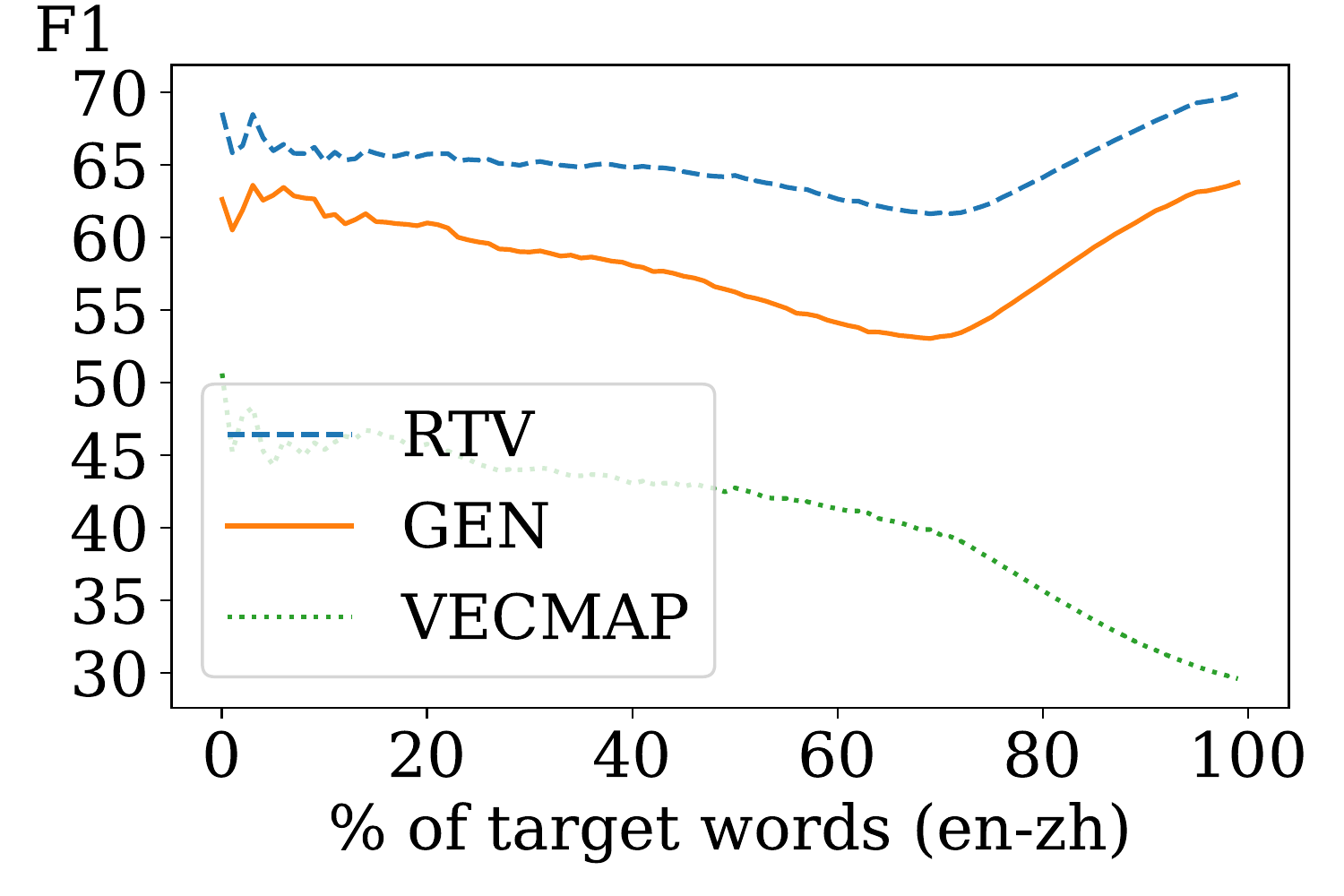}
\includegraphics[width=0.23\textwidth]{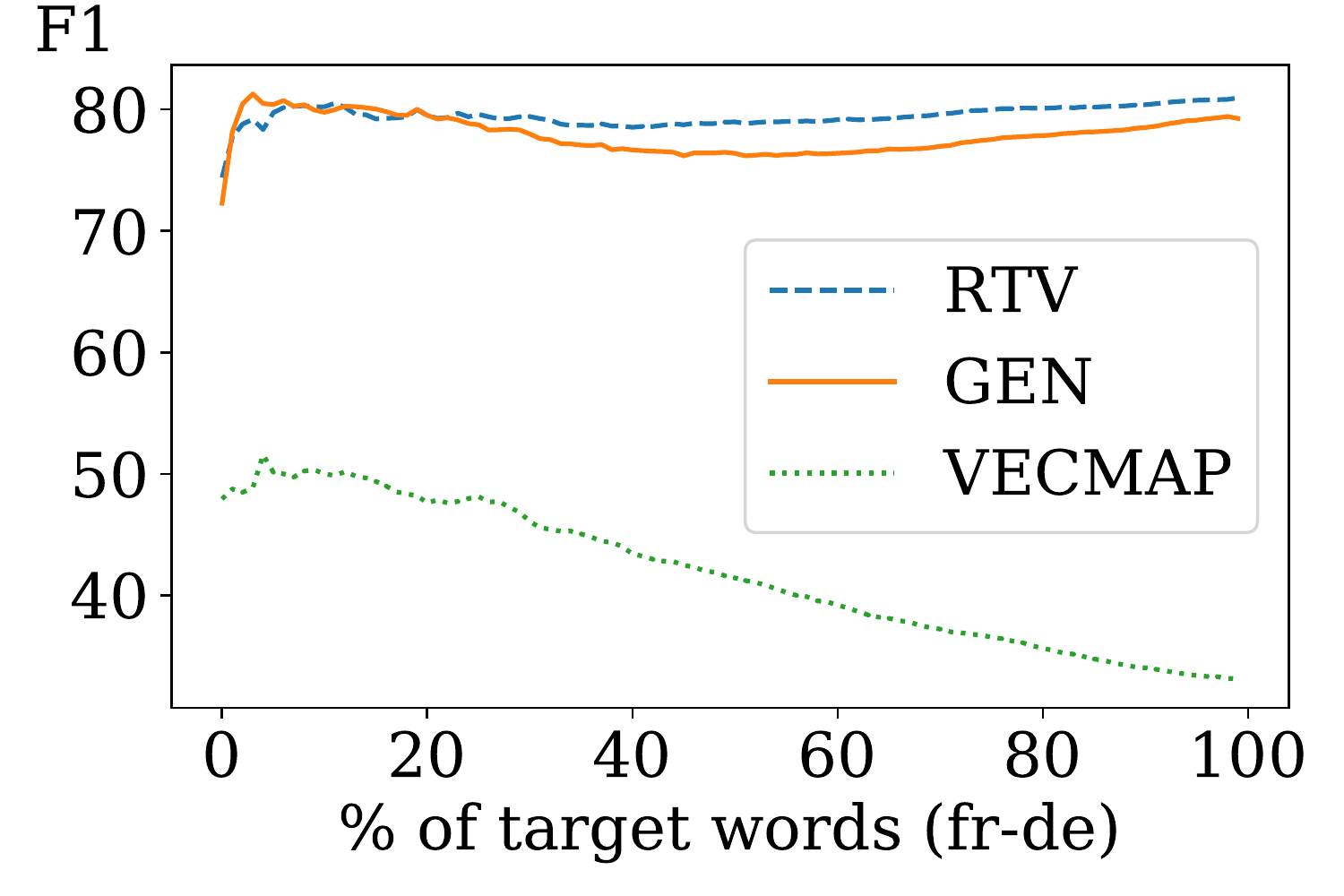} \\
\includegraphics[width=0.23\textwidth]{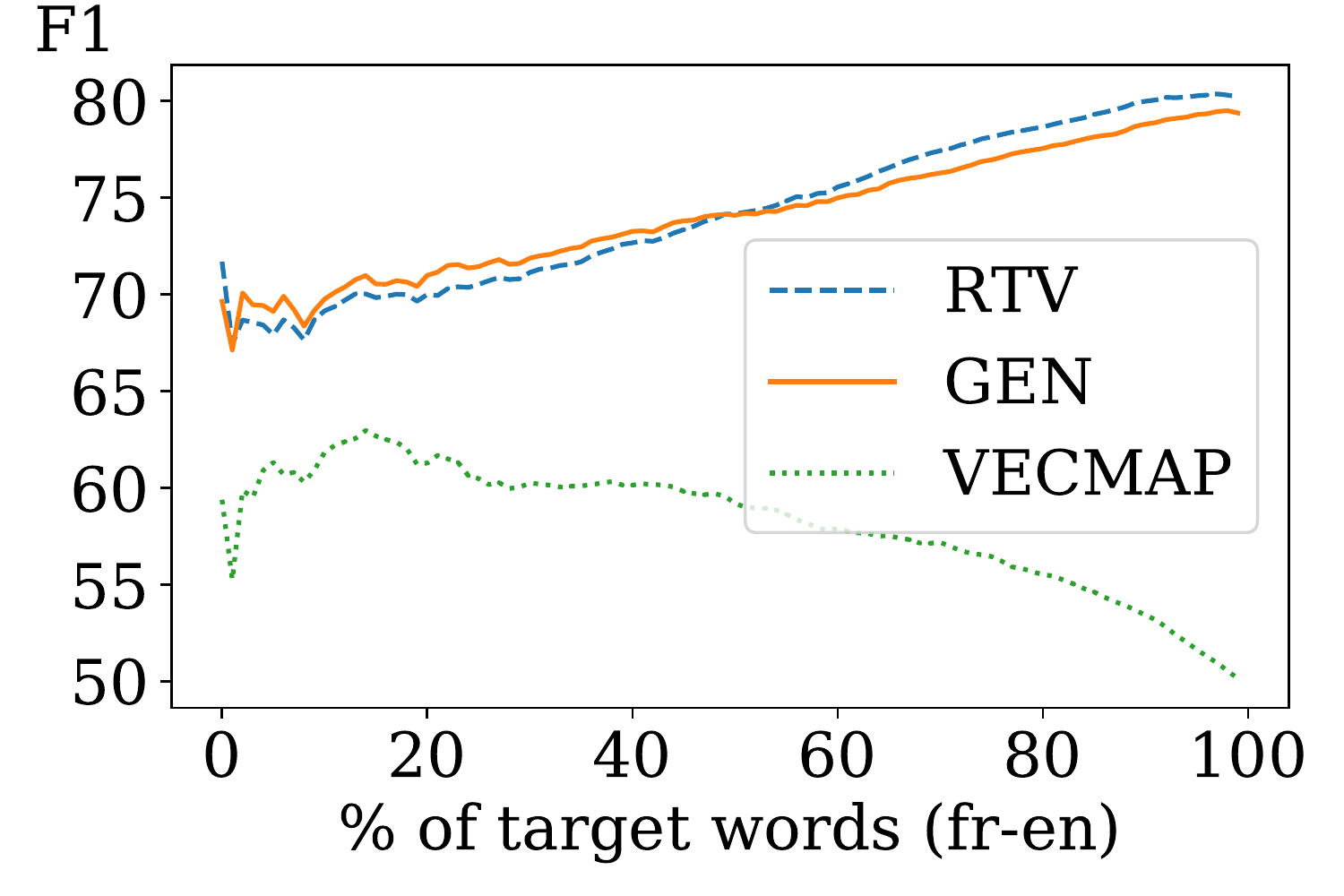}
\includegraphics[width=0.23\textwidth]{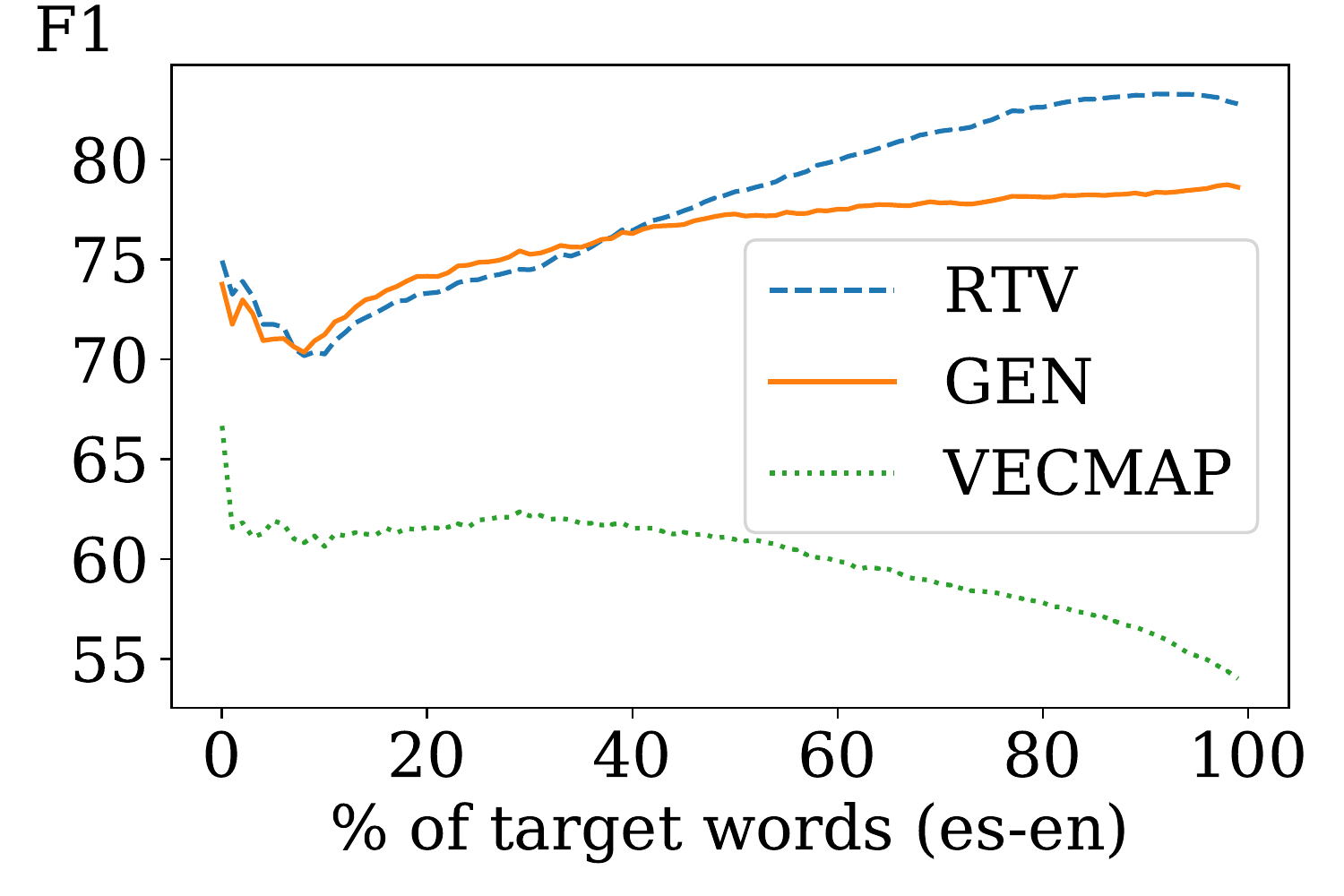}
\includegraphics[width=0.23\textwidth]{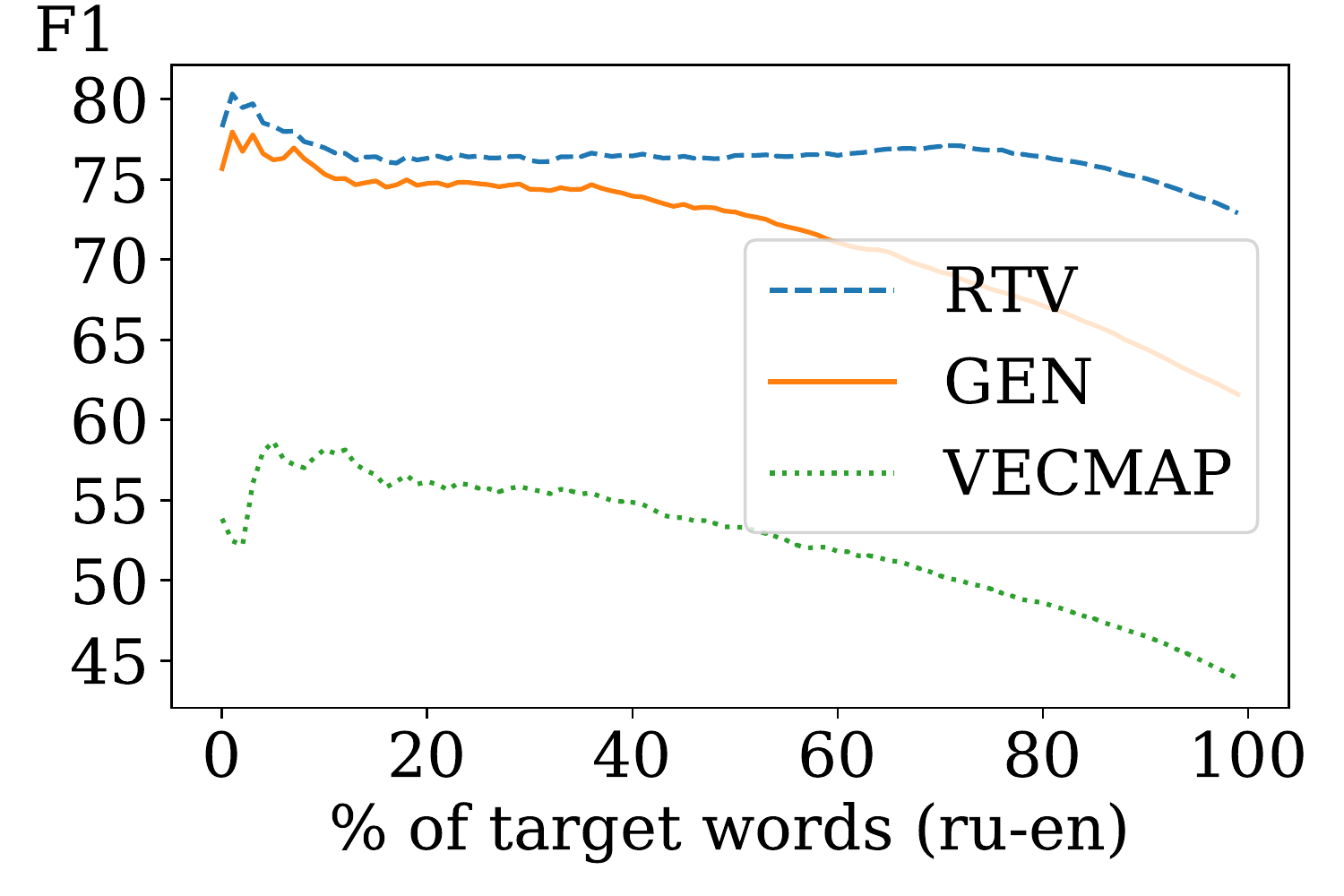}
\includegraphics[width=0.23\textwidth]{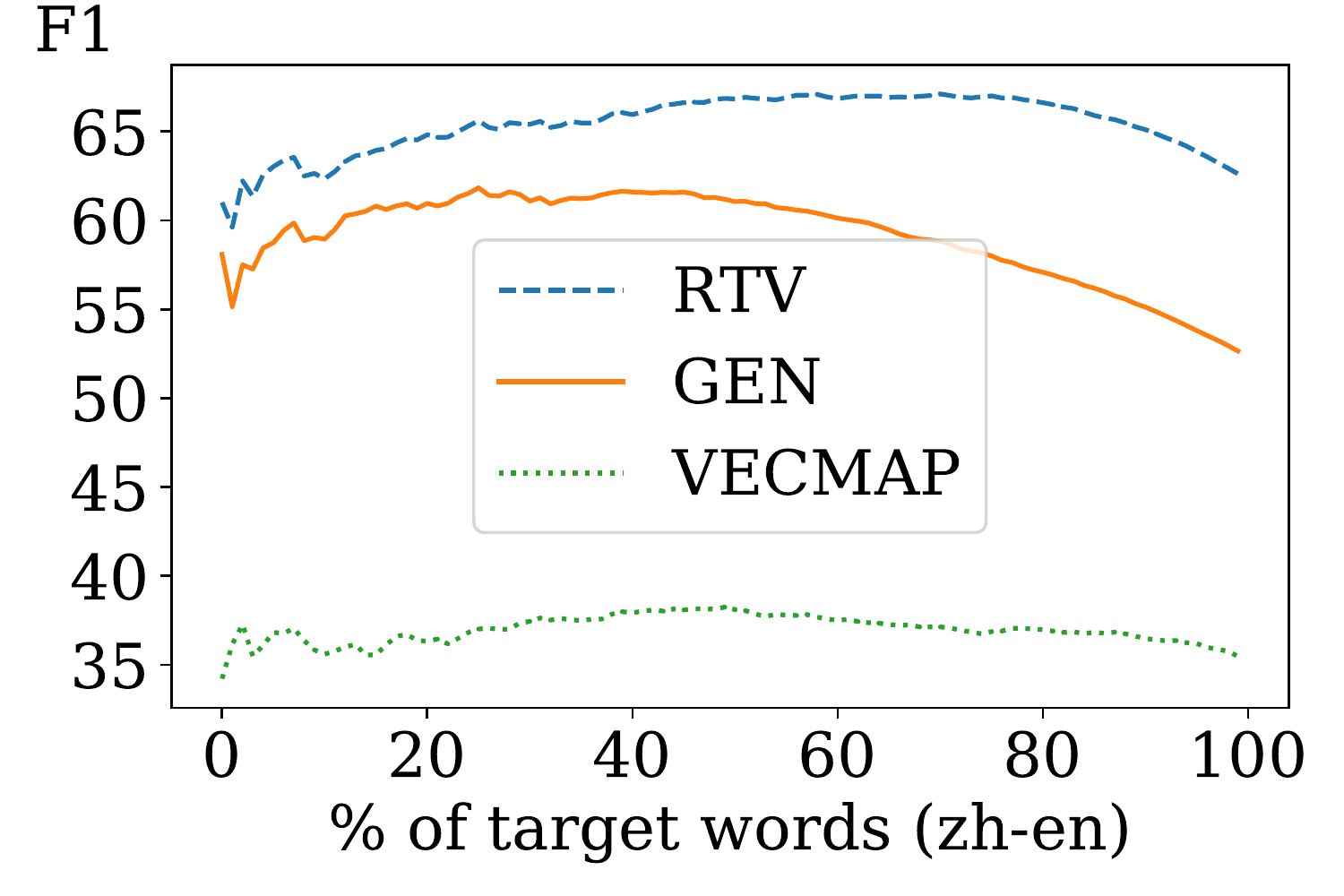}
\caption{\label{fig:lang-specific-trg} $F_1$ scores with respect to portion of target words kept for each investigated language pair, analogous to Figure~\ref{fig:tgt}.}
\end{figure*}

While Figure~\ref{fig:gen-rtv-src-freq} shows the average trend of $F_1$ scores with respect to the portion of source words or target words kept, we present such plots for each language pair in Figure~\ref{fig:lang-specific-src} and \ref{fig:lang-specific-trg}.
The trend of each separate method is inconsistent, which is consistent to the findings by BUCC 2020 participants \citep{rapp-etal-2020-overview}.
However, the conclusion that \textsc{rtv} gains more from low-frequency words still holds for most language pairs.

\section{Acceptability Judgments for en $\rightarrow$ zh}
\label{sec:error-cases-enzh}
\begin{table}[H]
    \centering \small 
    \begin{tabular}{lc|lc}
    \toprule
    \multicolumn{2}{c|}{\textsc{gen-rtv}} & \multicolumn{2}{c}{\textsc{VecMap}}  \\
    \midrule 
        southwestern  \hfill 西南部	&	\cmark 	&	spiritism \hfill      扶箕 & \xmark \\
        subject  \hfill  話題	&	\cmark 	    &	danny \hfill  john & \xmark \\
        screenwriter  \hfill  劇作家	&	\bf ?	&	hubbard \hfill        威廉斯 & \xmark \\
        preschool  \hfill  學齡前	&	\cmark 	&	swizz \hfill  incredible & \xmark  \\
        palestine  \hfill  palestine	&	\xmark 	&	viewing \hfill        觀賞 & \bf ? \\
        strengthening  \hfill  強化	&	\cmark 	&	prohibition \hfill    禁令 & \cmark \\
        zero  \hfill  0	&	\bf \cmark 	&	tons \hfill   滿載 & \xmark \\
        insurance  \hfill  保險公司	&	\xmark 	&	pascal \hfill         帕斯卡 & \cmark \\
        lines  \hfill  線路	&	\cmark 	&	claudia \hfill        christina & \xmark \\
        suburban  \hfill  市郊	&	\cmark 	&	massive \hfill       巨大 & \cmark \\
        honorable  \hfill  尊貴	&	\bf ? 	&	equity \hfill        估值 & \xmark \\
        placement  \hfill  置入	&	\cmark 	&	sandy \hfill  沙質 & \cmark  \\
        lesotho  \hfill  萊索托	&	\cmark 	&	fwd \hfill    不過後 & \xmark  \\
        shanxi  \hfill  shanxi	&	\xmark 	&	taillight \hfill      煞車燈 & \bf ? \\
        registration  \hfill  注冊	&	\cmark 	&	horoscope~ \hfill      生辰八字 & \xmark \\
        protestors  \hfill  抗議者	&	\cmark 	&	busan \hfill  仁川 & \xmark \\
        shovel  \hfill 剷	&	\cmark 	&	hiding \hfill         躲藏 & \cmark \\
        side  \hfill  一方	&	\cmark 	&	entry   \hfill 關時 & \xmark \\
        turbulence  \hfill  湍流	&	\cmark 	& weekends        \hfill 雙休日 & \bf ? \\
        omnibus \hfill omnibus	&	\xmark 	&	flagbearer      \hfill 掌旗 & \cmark \\
    \bottomrule 
    \end{tabular}
    \caption{Manually labeled acceptability judgments for random 20 error cases in English to Chinese translation made by \textsc{gen-rtv} and \textsc{VecMap}. }
    \label{tab:error-cases-enzh}
\end{table}

We present error analysis for the induced lexicon for English to Chinese translations (Table~\ref{tab:error-cases-enzh}) using the same method as Table~\ref{tab:error-cases}.
In this direction, many of the unacceptable cases are copying English words as their Chinese translations, which is also observed by \citet{rapp-etal-2020-overview}. This is due to an idiosyncrasy of the evaluation data where many English words are considered acceptable Chinese translations of the same words.

\section{Examples for Bitext in Different Sections}
\label{sec:qualitytiers}
We show examples of mined bitext with different quality (Table~\ref{tab:bitext-example}), where the mined bitexts are divided into 5 sections with respect to the similarity-based margin score (Eq~\ref{eq:margin-score}). 
The Chinese sentences are automatically converted to traditional Chinese alphabets using {\tt chinese\_converter},\footnote{\url{https://pypi.org/project/chinese-converter/}} to keep consistent with the MUSE dataset.

Based on our knowledge about these languages, we see that the \textsc{rtv-1} mostly consists of correct translations. 
While the other sections of bitext are of less quality, sentences within a pair are highly related or can be even partially aligned; therefore our bitext mining and alignment framework can still extract high-quality lexicon from such imperfect bitext. 
\begin{table*}[!ht]
    \centering \small 
    \begin{tabular}{c|l}
    \toprule 
         zh-en &  許多自然的問題實際上是承諾問題 。 \hfill  Many natural problems are actually promise problems. \\
         \sc rtv-1 &  寒冷氣候可能會帶來特殊挑戰。 \hfill  Cold climates may present special challenges. \\
         & 很顯然,曾經在某個場合達成了其所不知道的某種協議。\hfill I thought they'd come to some kind of an agreement. \\
         & 劇情發展順序與原作漫畫有些不同。 \hfill The plotline is somewhat different from the first series. \\
         & 他也創作過油畫和壁畫。 \hfill He also made sketches and paintings. \\
    \midrule 
         zh-en & 此節目被批評為宣揚偽科學和野史。\hfill  The book was criticized for misrepresenting nutritional science. \\
         \sc rtv-2 & 威藍町體育運動場 \hfill  Kawagoe Sports Park Athletics Stadium \\
         & 他是她的神聖醫師和保護者。 \hfill He's her protector and her provider. \\
         & 其後以5,000英鎊轉會到盧頓。\hfill He later returned to Morton for £15,000. \\
         & 滬生和阿寶是小說的兩個主要人物。\hfill Lawrence and Joanna are the play's two major characters. \\
    \midrule 
         zh-en & 一般上沒有會員加入到母政黨。 \hfill  Voters do not register as members of political parties. \\
         \sc rtv-3 & 曾任《紐約時報》書評人。 \hfill  He was formerly an editor of ``The New York Times Book Review'' . \\
         & 48V微混系統主要由以下組件構成: \hfill  The M120 mortar system consists of the following major components: \\
         & 其後以5,000英鎊轉會到盧頓。\hfill He later returned to Morton for £15,000. \\
         & 2月25日從香港抵達汕頭 \hfill  and arrived at Hobart Town on 8 November. \\
    \midrule 
        zh-en &  1261年,拉丁帝國被推翻,東羅馬帝國復國。 \hfill  The Byzantine Empire was fully reestablished in 1261. \\
        \sc rtv-4 & 而這次航行也證明他的指責是正確的。 \hfill  This proved that he was clearly innocent of the charges.\\
        & 並已經放出截面和試用版。 \hfill  A cut-down version was made available for downloading. \\
        & 它重370克,由一根把和九根索組成。 \hfill  It consists of 21 large gears and a 13 meters pendulum. \\
        & 派路在隊中的創造力可謂無出其右,功不可抹。 \hfill  Still, the German performance was not flawless. \\
    \midrule 
        zh-en &  此要塞也用以鎮壓的部落。 \hfill  that were used by nomads in the region. \\
        \sc rtv-5 & 不過,這31次出場只有11次是首發。 \hfill  In those 18 games, the visiting team won only three times.\\
        & 生於美國紐約州布魯克林。 \hfill  He was born in Frewsburg, New York, USA.\\
        & 2014年7月14日,組團成為一員。 \hfill  Roy joined the group on 4/18/98. \\
        & 盾上有奔走中的獅子。 \hfill  Far above, the lonely hawk floating. \\
    \midrule \midrule 
        de-en & Von 1988 bis 1991 lebte er in Venedig.  \hfill  From 1988-1991 he lived in Venice. \\
        \sc rtv-1 &  Der Film beginnt mit folgendem Zitat:  \hfill  The movie begins with the following statement: \\
        & Geschichte von Saint Vincent und den Grenadinen  \hfill   History of Saint Vincent and the Grenadines \\
        & Die Spuren des Kriegs sind noch allgegenwärtig.  \hfill  Some signs of the people are still there. \\
        & Saint-Paul (Savoie)  \hfill  Saint-Paul, Savoie \\
    \midrule
        de-en & Nanderbarsche sind nicht brutpflegend.  \hfill  Oxpeckers are fairly gregarious. \\
        \sc rtv-2 & Dort begegnet sie Raymond und seiner Tochter Sarah.  \hfill  There she meets Sara and her husband. \\
        & Armansperg wurde zum Premierminister ernannt.  \hfill  Mansur was appointed the prime minister. \\
        & Diese Arbeit wird von den Männchen ausgeführt.  \hfill  Parental care is performed by males. \\
        & August von Limburg-Stirum  \hfill  House of Limburg-Stirum \\
    \midrule 
        de-en & Es gibt mehrere Anbieter der Komponenten.  \hfill  There are several components to the site.\\
        \sc rtv-3 & Doch dann werden sie von Piraten angegriffen.  \hfill  They are attacked by Saracen pirates.\\
        & Wird nicht die tiefste – also meist 6.  \hfill  The shortest, probably five.\\
        & Ihre Blüte hatte sie zwischen 1976 und 1981.  \hfill  The crop trebled between 1955 and 1996.\\
        & Er brachte Reliquien von der Hl.  \hfill  Eulogies were given by the Rev.\\
     \midrule 
        de-en & Gespielt wird meistens Mitte Juni.  \hfill  It is played principally on weekends. \\
        \sc rtv-4 & Schuppiger Schlangenstern  \hfill  Plains garter snake\\
        & Das Artwork stammt von Dave Field.  \hfill  The artwork is by Mike Egan.\\
        & Ammonolyse ist eine der Hydrolyse analoge Reaktion,  \hfill  Hydroxylation is an oxidative process. \\
        & Die Pellenz gliedert sich wie folgt:  \hfill  The Pellenz is divided as follows:\\
     \midrule 
        de-en & Auch Nicolau war praktizierender Katholik.  \hfill  Cassar was a practicing Roman Catholic.\\
        \sc rtv-5 & Im Jahr 2018 lag die Mitgliederzahl bei 350.  \hfill  The membership in 2017 numbered around 1,000.\\
        & Er trägt die Fahrgestellnummer TNT 102.  \hfill  It carries the registration number AWK 230.\\
        & Als Moderator war Benjamin Jaworskyj angereist.  \hfill  Dmitry Nagiev appeared as the presenter.\\
        & Benachbarte Naturräume und Landschaften sind:  \hfill  Neighboring hydrographic watersheds are:\\
    \bottomrule 
    \end{tabular}
    \caption{Examples of bitext in different sections (Section~\ref{sec:bitext-quality}). We see that tier 1 has majority parallel sentences whereas lower tiers have mostly similar but not parallel sentences. }
    \label{tab:bitext-example}
\end{table*}

\section{Results: P@1 on the MUSE Dataset}
\label{sec:p1eval}
Precision@1 (P@1) is a widely applied metric to evaluate bilingual lexicon induction \citep[\emph{inter alia}]{smith2017offline,conneau2017word,artetxe-etal-2019-bilingual}, therefore we compare our models with existing approaches in terms of P@1 as well (Table~\ref{tab:precision-fully-unsup}). Our fully unsupervised method with retrieval-based bitext outperforms the previous state of the art \citep{artetxe-etal-2019-bilingual} by 4.1 average P@1, and achieve competitive or superior performance on all investigated language pairs. 

\begin{table*}[!ht]
\begin{center}
\begin{small}
  \addtolength{\tabcolsep}{-1pt}
  \begin{tabular}{lcccccccccccccc}
    \toprule
    && \multicolumn{2}{c}{en-es} && \multicolumn{2}{c}{en-fr} && \multicolumn{2}{c}{en-de} && \multicolumn{2}{c}{en-ru} && \multirow{2}{*}{avg.} \\
    \cmidrule{3-4} \cmidrule{6-7} \cmidrule{9-10} \cmidrule{12-13}
    && $\rightarrow$ & $\leftarrow$ && $\rightarrow$ & $\leftarrow$ && $\rightarrow$ & $\leftarrow$ && $\rightarrow$ & $\leftarrow$ && \\
    \midrule
    Nearest neighbor$^\dagger$ && 81.9 & 82.8 && 81.6 & 81.7 && 73.3 & 72.3 && 44.3 & 65.6 && 72.9 \\
    Inv. nearest neighbor \citep{dinu2015improving}$^\dagger$ && 80.6 & 77.6 && 81.3 & 79.0 && 69.8 & 69.7 && 43.7 & 54.1 && 69.5 \\
    Inv. softmax \citep{smith2017offline}$^\dagger$ && 81.7 & 82.7 && 81.7 & 81.7 && 73.5 & 72.3 && 44.4 & 65.5 && 72.9 \\
    CSLS \citep{conneau2017word}$^\dagger$ && 82.5 & 84.7 && 83.3 & 83.4 && 75.6 & 75.3 && 47.4 & 67.2 && 74.9 \\
    \midrule 
    \citet{artetxe-etal-2019-bilingual}$^\dagger$ && 87.0 & 87.9 && \bf 86.0 & 86.2 && 81.9 & 80.2 && 50.4 & 71.3 && 78.9 \\
    \midrule 
    \textsc{rtv} (ours) && \bf 89.9 & \bf 93.5 && 84.5 & \bf 89.5 && \bf 83.0 & \bf 88.6 && \bf 54.5 & \bf 80.7 && \bf 83.0 \\
    \textsc{gen} (ours) && 81.5 & 88.7 && 81.6 & 88.6 && 78.9 & 83.7 && 35.4 & 68.2 && 75.8 \\
    \bottomrule
  \end{tabular}
\end{small}
\end{center}
\caption{ P@1 of our lexicon inducer and previous methods on the standard MUSE test set \citep{conneau2017word}, where the best number in each column is bolded.  The first section consists of vector rotation--based methods, while \citet{artetxe-etal-2019-bilingual} conduct unsupervised machine translation and word alignment to induce bilingual lexicons. All methods are tested in the fully unsupervised setting. $\dagger$: numbers copied from \citet{artetxe-etal-2019-bilingual}. } 
\label{tab:precision-fully-unsup}
\end{table*}

\section{Error analysis}
\label{sec:realerrors}
To understand the remaining errors, we randomly sampled 400 word pairs from the induced lexicon and compare them to ground truth as and Google Translate via \texttt{=googletranslate(A1, "zh", "en")}.
All error cases are included in Table~\ref{tab:erroranalysis}.
In overall precision, our induced lexicon is comparable to the output of Google translate API where
there are 17 errors for \textsc{gen-rtv}
14 errors for Google and 4 common errors.

\begin{table*}[!ht]
\centering
\begin{small}
\begin{tabular}{llll}
\toprule
src & \textsc{gen-rtv}          &             & Google Trans.            \\
\midrule
編劇  & writers       & \textless{}    & screenwriter      \\
可笑  & laughing      & \textless{}    & ridiculous        \\
極權  & authoritarian & \textless{}    & Totalitarian      \\
押韻  & couplets      & \textless{}    & rhyme             \\
烙印  & tattooed      & \textless{}    & brand             \\
業主  & homeowners    & \textless{}    & owner             \\
安娜  & grande        & \textless{}    & Anna              \\
包頭  & header        & \textless{}    & Baotou            \\
編輯  & editorial     & \textless{}    & edit              \\
陣風  & winds         & \textless{}    & gust              \\
火柴  & firewood      & \textless{}    & matches           \\
盃   & bowl          & \textless{}    & cup               \\
武士道 & samurai       & \textless{}    & Bushido           \\
詩句  & poem          & \textless{}    & verse             \\
肚臍  & belly         & \textless{}    & belly button      \\
現代化 & modern        & \textless{}    & modernization     \\
感冒  & flu           & \textless{}    & cold              \\
\midrule
協商  & negotiate     & \textgreater{} & Consult           \\
納米  & nanometer     & \textgreater{} & Nano              \\
類人猿 & apes          & \textgreater{} & Anthropoid        \\
配件  & accessories   & \textgreater{} & Fitting           \\
匯   & aggregated    & \textgreater{} & exchange          \\
貸方  & lenders       & \textgreater{} & Credit            \\
逆差  & deficit       & \textgreater{} & Trade deficit     \\
如果  & if            & \textgreater{} & in case           \\
附件  & accessories   & \textgreater{} & annex             \\
實習  & internship    & \textgreater{} & practice          \\
加冕  & crowned       & \textgreater{} & Crown             \\
助理  & assistant     & \textgreater{} & assistant Manager \\
親和性 & agreeableness & \textgreater{} & Affinity          \\
國土  & homeland      & $>$ & land              \\
\midrule
過境  & crossings     & \xmark             & Transit           \\
環流  & circulation   & \xmark              & Circumfluence     \\
羊群  & sheep         & \xmark              & Herd \\           \bottomrule
\end{tabular}
\end{small}
\caption{all errors cases among 400 random outputs of \textsc{gen-rtv}
compared to both our judgement and Google translate for reference.
$>$: \textsc{gen-rtv} unacceptable while Google Trans acceptable.
$<$: \textsc{gen-rtv} acceptable while Google Trans unacceptable.
\xmark{}: both unacceptable.
}
\label{tab:erroranalysis}
\end{table*}

\clearpage
\section{Comparison between MUSE and \textsc{gen-rtv}}
\label{appendix:manual-judgments}
We present the word pairs involved in the comparison between MUSE benchmark and our \textsc{gen-rtv} method in Table~\ref{tab:muse-genrtv-judgments}. 

{
\small \centering
\topcaption{Word pairs from the union of MUSE and \textsc{gen-rtv} lexicon and manual judgments. M: entries from MUSE, G: entries from the \textsc{gen-rtv} lexicon, B: entries from both sources. \label{tab:muse-genrtv-judgments}}
\tablefirsthead{\toprule zh	&	en	&\bf	Src.	&	\bf Acc. \\ \midrule}
\tablehead{%
\multicolumn{4}{c}{{\bfseries  Table~\ref{tab:muse-genrtv-judgments}, Continued from previous column}} \\
\toprule zh	&	en	&\bf	Src.	&	\bf Acc. \\ 
\midrule
}
\tabletail{
\midrule \multicolumn{4}{r}{{Continued on next column}} \\ \midrule}
\tablelasttail{%
\\\midrule
\multicolumn{4}{r}{{End of Table.}} \\ \bottomrule}
\begin{xtabular}{lrcc}
$\geq$	&	$\geq$	&	B	&	\cmark	\\
中共	&	cpc	&	M	&	\cmark	\\
中共	&	ccp	&	M	&	\cmark	\\
中西	&	midwest	&	M	&	\cmark	\\
亞洲	&	asia	&	B	&	\cmark	\\
亞洲	&	asian	&	B	&	\cmark	\\
交換	&	exchanged	&	G	&	\cmark	\\
交換	&	exchanging	&	G	&	\cmark	\\
交換	&	exchange	&	B	&	\cmark	\\
交換	&	swap	&	B	&	\cmark	\\
仇恨	&	hate	&	B	&	\cmark	\\
仇恨	&	hatred	&	B	&	\cmark	\\
伊朗	&	iran	&	B	&	\cmark	\\
伊朗	&	iranian	&	B	&	\cmark	\\
估算	&	estimates	&	G	&	\cmark	\\
估算	&	estimate	&	G	&	\cmark	\\
估算	&	estimation	&	M	&	\cmark	\\
估算	&	estimating	&	G	&	\cmark	\\
估算	&	estimated	&	G	&	\cmark	\\
佔有	&	possession	&	M	&	\cmark	\\
佛教徒	&	buddhist	&	B	&	\cmark	\\
佛教徒	&	buddhists	&	B	&	\cmark	\\
依偎	&	snuggle	&	M	&	\cmark	\\
依然	&	still	&	B	&	\cmark	\\
修正	&	amendment	&	M	&	\cmark	\\
修正	&	corrected	&	G	&	\cmark	\\
修正	&	amendments	&	M	&	\cmark	\\
修正	&	amend	&	M	&	\cmark	\\
修正	&	corrections	&	G	&	\cmark	\\
俾斯麥	&	bismarck	&	B	&	\cmark	\\
借調	&	secondment	&	M	&	\cmark	\\
借調	&	seconded	&	B	&	\cmark	\\
停車場	&	parking	&	B	&	\xmark	\\
傻	&	stupid	&	B	&	\cmark	\\
傻	&	silly	&	M	&	\cmark	\\
光澤	&	shiny	&	G	&	\cmark	\\
光澤	&	gloss	&	M	&	\cmark	\\
公羊	&	ram	&	B	&	\cmark	\\
公羊	&	rams	&	M	&	\cmark	\\
凸顯	&	highlight	&	M	&	\cmark	\\
出遊	&	outing	&	M	&	\cmark	\\
出遊	&	outings	&	M	&	\cmark	\\
前奏	&	prelude	&	B	&	\cmark	\\
前奏	&	foreplay	&	M	&	\xmark	\\
努	&	nu	&	M	&	\xmark	\\
包租	&	charters	&	M	&	\cmark	\\
包租	&	charter	&	M	&	\cmark	\\
包租	&	chartered	&	M	&	\cmark	\\
包租	&	chartering	&	M	&	\cmark	\\
匈牙利語	&	hungarian	&	B	&	\cmark	\\
匯款	&	remittance	&	B	&	\cmark	\\
匯款	&	remittances	&	B	&	\cmark	\\
協約	&	concordat	&	M	&	\cmark	\\
協約	&	pact	&	M	&	\cmark	\\
協約	&	compact	&	M	&	\cmark	\\
原木	&	logs	&	B	&	\cmark	\\
參考	&	reference	&	B	&	\cmark	\\
句法	&	syntactic	&	G	&	\cmark	\\
句法	&	syntax	&	B	&	\cmark	\\
同義字	&	synonym	&	B	&	\cmark	\\
同義字	&	synonyms	&	M	&	\cmark	\\
吱吱	&	zee	&	M	&	\xmark	\\
吱吱	&	squeaking	&	M	&	\cmark	\\
呼吸道	&	respiratory	&	G	&	\cmark	\\
呼吸道	&	airway	&	B	&	\cmark	\\
命名	&	named	&	G	&	\cmark	\\
命名	&	naming	&	B	&	\cmark	\\
咖哩	&	curry	&	M	&	\cmark	\\
哈爾濱	&	harbin	&	B	&	\cmark	\\
問候	&	greeting	&	B	&	\cmark	\\
問候	&	greetings	&	B	&	\cmark	\\
喇叭	&	horns	&	G	&	\cmark	\\
喇叭	&	horn	&	M	&	\cmark	\\
嘔吐	&	vomited	&	G	&	\cmark	\\
嘔吐	&	vomit	&	G	&	\cmark	\\
嘔吐	&	vomiting	&	B	&	\cmark	\\
嚴酷	&	harsh	&	B	&	\cmark	\\
因此	&	so	&	M	&	\cmark	\\
因此	&	hence	&	G	&	\cmark	\\
因此	&	therefore	&	G	&	\cmark	\\
因此	&	thus	&	G	&	\cmark	\\
國家	&	state	&	M	&	\cmark	\\
國家	&	countries	&	B	&	\cmark	\\
國家	&	country	&	B	&	\cmark	\\
國家	&	national	&	B	&	\cmark	\\
國家	&	states	&	M	&	\cmark	\\
地主	&	landowners	&	B	&	\cmark	\\
地主	&	landlords	&	G	&	\cmark	\\
地主	&	landlord	&	B	&	\cmark	\\
地熱	&	geothermal	&	B	&	\cmark	\\
地鐵	&	metro	&	B	&	\cmark	\\
地鐵	&	subway	&	B	&	\cmark	\\
增長	&	growing	&	G	&	\cmark	\\
增長	&	growth	&	B	&	\cmark	\\
多雲	&	cloudy	&	B	&	\cmark	\\
夜總會	&	nightclub	&	B	&	\cmark	\\
夜總會	&	nightclubs	&	B	&	\cmark	\\
大島	&	oshima	&	M	&	\cmark	\\
奴隸制	&	slavery	&	B	&	\cmark	\\
姐妹	&	sister	&	B	&	\cmark	\\
姐妹	&	sisters	&	B	&	\cmark	\\
媽祖	&	mazu	&	M	&	\cmark	\\
嫉妒	&	jealousy	&	B	&	\cmark	\\
嫉妒	&	envy	&	G	&	\cmark	\\
嫉妒	&	jealous	&	B	&	\cmark	\\
字面上	&	literally	&	B	&	\cmark	\\
學位	&	degrees	&	B	&	\cmark	\\
學位	&	degree	&	B	&	\cmark	\\
安培	&	ampere	&	G	&	\cmark	\\
安培	&	amperes	&	G	&	\cmark	\\
安培	&	amber	&	M	&	\xmark	\\
安迪	&	andy	&	B	&	\cmark	\\
官邸	&	residence	&	G	&	\xmark	\\
官邸	&	mansion	&	G	&	\xmark	\\
官邸	&	residences	&	M	&	\xmark	\\
客艙	&	cabin	&	B	&	\cmark	\\
客艙	&	cabins	&	G	&	\cmark	\\
家庭教師	&	tutor	&	G	&	\cmark	\\
家庭教師	&	governess	&	B	&	\cmark	\\
容器	&	container	&	B	&	\cmark	\\
容器	&	containers	&	B	&	\cmark	\\
寄生蟲	&	parasites	&	B	&	\cmark	\\
寄生蟲	&	parasite	&	B	&	\cmark	\\
寬大	&	leniency	&	M	&	\cmark	\\
寬大	&	clemency	&	M	&	\cmark	\\
專利權	&	patents	&	B	&	\cmark	\\
專利權	&	patent	&	B	&	\cmark	\\
小型	&	small	&	B	&	\cmark	\\
小型	&	smaller	&	G	&	\cmark	\\
小指	&	pinky	&	M	&	\cmark	\\
小行星	&	asteroid	&	B	&	\cmark	\\
小行星	&	asteroids	&	B	&	\cmark	\\
層層	&	layers	&	B	&	\cmark	\\
巡邏	&	patrol	&	B	&	\cmark	\\
巡邏	&	patrols	&	B	&	\cmark	\\
巡邏	&	patrolled	&	G	&	\cmark	\\
巡邏	&	patrolling	&	B	&	\cmark	\\
市鎮	&	municipality	&	G	&	\cmark	\\
市鎮	&	communes	&	M	&	\cmark	\\
市鎮	&	municipalities	&	B	&	\cmark	\\
幻滅	&	disillusionment	&	B	&	\cmark	\\
幾十年	&	decades	&	B	&	\cmark	\\
引用	&	references	&	G	&	\cmark	\\
引用	&	quote	&	M	&	\cmark	\\
引用	&	quoted	&	M	&	\cmark	\\
引用	&	quotes	&	M	&	\cmark	\\
引用	&	cited	&	G	&	\cmark	\\
引用	&	cite	&	G	&	\cmark	\\
引用	&	reference	&	G	&	\cmark	\\
彈性	&	flexibility	&	B	&	\cmark	\\
彈性	&	flexible	&	M	&	\cmark	\\
彈性	&	resilient	&	G	&	\cmark	\\
彈性	&	flex	&	M	&	\cmark	\\
彈性	&	elasticity	&	B	&	\cmark	\\
彈性	&	elastic	&	B	&	\cmark	\\
形式	&	forms	&	G	&	\cmark	\\
形式	&	format	&	M	&	\cmark	\\
形式	&	form	&	B	&	\cmark	\\
影像	&	images	&	B	&	\cmark	\\
影像	&	imaging	&	B	&	\cmark	\\
影像	&	image	&	B	&	\cmark	\\
往下	&	down	&	M	&	\xmark	\\
復活節	&	easter	&	B	&	\cmark	\\
必需品	&	essentials	&	B	&	\cmark	\\
必需品	&	necessities	&	B	&	\cmark	\\
必需品	&	necessity	&	G	&	\cmark	\\
懷舊	&	nostalgia	&	B	&	\cmark	\\
懷舊	&	nostalgic	&	B	&	\cmark	\\
懷舊	&	throwback	&	M	&	\cmark	\\
所有者	&	owners	&	B	&	\cmark	\\
所有者	&	owner	&	B	&	\cmark	\\
托兒所	&	nursery	&	B	&	\cmark	\\
托兒所	&	crèche	&	M	&	\cmark	\\
托兒所	&	daycare	&	M	&	\cmark	\\
托兒所	&	nurseries	&	M	&	\cmark	\\
托盤	&	tray	&	B	&	\cmark	\\
托盤	&	trays	&	B	&	\cmark	\\
托盤	&	pallets	&	G	&	\cmark	\\
扣	&	buckle	&	M	&	\cmark	\\
投影	&	projective	&	G	&	\cmark	\\
投影	&	projected	&	G	&	\cmark	\\
投影	&	projections	&	G	&	\cmark	\\
投影	&	projection	&	B	&	\cmark	\\
抽籤	&	draw	&	B	&	\xmark	\\
拉鏈	&	zipper	&	M	&	\cmark	\\
拉鏈	&	zippers	&	G	&	\cmark	\\
拉麵	&	ramen	&	M	&	\cmark	\\
接吻	&	kissing	&	B	&	\cmark	\\
接吻	&	kiss	&	G	&	\cmark	\\
日本人	&	japanese	&	M	&	\cmark	\\
晶格	&	lattice	&	B	&	\cmark	\\
暴發	&	outbreak	&	M	&	\cmark	\\
暴發	&	outbreaks	&	G	&	\cmark	\\
曆	&	calendars	&	G	&	\cmark	\\
曆	&	calendar	&	B	&	\cmark	\\
曲軸	&	crankshaft	&	B	&	\cmark	\\
曼谷	&	bangkok	&	B	&	\cmark	\\
極	&	polar	&	M	&	\cmark	\\
極	&	extremely	&	B	&	\cmark	\\
樂器	&	instrument	&	B	&	\cmark	\\
樂器	&	instruments	&	B	&	\cmark	\\
樹枝	&	branches	&	M	&	\cmark	\\
橋梁	&	bridges	&	B	&	\cmark	\\
橋梁	&	bridge	&	B	&	\cmark	\\
機遇	&	opportunities	&	B	&	\cmark	\\
機遇	&	opportunity	&	B	&	\cmark	\\
橫濱	&	yokohama	&	M	&	\cmark	\\
比薩	&	pizza	&	M	&	\cmark	\\
比薩	&	pisa	&	G	&	\cmark	\\
氣動	&	pneumatic	&	B	&	\cmark	\\
氣動	&	aerodynamic	&	G	&	\cmark	\\
決算	&	resolutions	&	G	&	\xmark	\\
決算	&	accounts	&	M	&	\xmark	\\
油漆	&	paints	&	B	&	\cmark	\\
油漆	&	paint	&	B	&	\cmark	\\
油菜	&	canola	&	M	&	\cmark	\\
泡沫	&	bubble	&	M	&	\cmark	\\
泡沫	&	foam	&	M	&	\cmark	\\
泡沫	&	bubbles	&	M	&	\cmark	\\
泥漿	&	muds	&	G	&	\cmark	\\
泥漿	&	mud	&	B	&	\cmark	\\
深不可測	&	unfathomable	&	M	&	\xmark	\\
湯頭	&	tonto	&	M	&	\xmark	\\
漆	&	varnish	&	M	&	\xmark	\\
漆	&	lacquer	&	M	&	\cmark	\\
漆	&	paint	&	G	&	\cmark	\\
潛力	&	potentials	&	M	&	\cmark	\\
潛力	&	potential	&	B	&	\cmark	\\
濕疹	&	eczema	&	B	&	\cmark	\\
火災	&	fires	&	B	&	\cmark	\\
火災	&	fire	&	B	&	\cmark	\\
烈	&	yeol	&	M	&	\xmark	\\
焦點	&	focus	&	B	&	\cmark	\\
照片	&	pictures	&	M	&	\cmark	\\
照片	&	photos	&	B	&	\cmark	\\
照片	&	picture	&	M	&	\cmark	\\
照片	&	photographs	&	B	&	\cmark	\\
照片	&	photograph	&	B	&	\cmark	\\
照片	&	photo	&	B	&	\cmark	\\
燒瓶	&	flask	&	B	&	\cmark	\\
牛角	&	horn	&	M	&	\xmark	\\
牛角	&	horns	&	M	&	\xmark	\\
牡丹	&	peony	&	M	&	\cmark	\\
玩	&	play	&	B	&	\cmark	\\
玩	&	playing	&	B	&	\cmark	\\
現任	&	present	&	M	&	\cmark	\\
現任	&	incumbent	&	G	&	\cmark	\\
現任	&	current	&	M	&	\cmark	\\
理想	&	ideals	&	B	&	\cmark	\\
理想	&	ideally	&	G	&	\cmark	\\
理想	&	ideal	&	B	&	\cmark	\\
田野	&	fields	&	M	&	\cmark	\\
白蘭地	&	brandy	&	B	&	\cmark	\\
百合	&	lily	&	M	&	\cmark	\\
百合	&	lilium	&	M	&	\cmark	\\
百合	&	lilies	&	M	&	\cmark	\\
監視	&	monitoring	&	G	&	\cmark	\\
監視	&	surveillance	&	B	&	\cmark	\\
監視	&	monitor	&	G	&	\cmark	\\
直立	&	upright	&	B	&	\cmark	\\
相似	&	similarity	&	M	&	\cmark	\\
相似	&	resemblance	&	M	&	\cmark	\\
相似	&	similar	&	B	&	\cmark	\\
省份	&	province	&	B	&	\cmark	\\
省份	&	provinces	&	B	&	\cmark	\\
眼皮	&	eyelid	&	M	&	\cmark	\\
眼皮	&	eyelids	&	B	&	\cmark	\\
瞬時	&	instantaneous	&	B	&	\cmark	\\
瞬時	&	instantaneously	&	G	&	\cmark	\\
祝願	&	wish	&	M	&	\cmark	\\
祝願	&	wishes	&	B	&	\cmark	\\
移交	&	handover	&	M	&	\cmark	\\
移交	&	transferred	&	G	&	\cmark	\\
篩選	&	filters	&	M	&	\cmark	\\
篩選	&	screening	&	B	&	\cmark	\\
篩選	&	filter	&	M	&	\cmark	\\
篩選	&	filtering	&	M	&	\cmark	\\
簽字	&	signature	&	M	&	\cmark	\\
簽字	&	signed	&	G	&	\cmark	\\
簽字	&	sign	&	B	&	\cmark	\\
籃球	&	basketball	&	B	&	\cmark	\\
紀念館	&	memorial	&	B	&	\cmark	\\
紀念館	&	memorials	&	G	&	\cmark	\\
紅寶石	&	ruby	&	B	&	\cmark	\\
紅寶石	&	rubies	&	M	&	\cmark	\\
納米	&	nano	&	M	&	\xmark	\\
納米	&	nm	&	G	&	\cmark	\\
納米	&	nanometer	&	G	&	\cmark	\\
納米	&	nanometers	&	G	&	\cmark	\\
素描	&	sketch	&	B	&	\cmark	\\
素描	&	sketching	&	M	&	\cmark	\\
素描	&	sketches	&	M	&	\cmark	\\
素描	&	drawings	&	G	&	\cmark	\\
綻放	&	blooming	&	M	&	\cmark	\\
綻放	&	blossom	&	G	&	\cmark	\\
綻放	&	bloom	&	B	&	\cmark	\\
綻放	&	blossoming	&	M	&	\cmark	\\
繁殖	&	reproduction	&	B	&	\cmark	\\
繁殖	&	breed	&	B	&	\cmark	\\
繁殖	&	breeding	&	B	&	\cmark	\\
繁殖	&	propagation	&	M	&	\cmark	\\
繁殖	&	reproduce	&	G	&	\cmark	\\
繩索	&	ropes	&	G	&	\cmark	\\
繩索	&	rope	&	B	&	\cmark	\\
罌粟	&	poppy	&	B	&	\cmark	\\
罌粟	&	poppies	&	M	&	\cmark	\\
義大利文	&	italian	&	B	&	\cmark	\\
肉汁	&	gravy	&	B	&	\cmark	\\
肚臍	&	belly	&	G	&	\xmark	\\
肚臍	&	navel	&	B	&	\cmark	\\
腎	&	kidney	&	B	&	\cmark	\\
腎	&	renal	&	B	&	\cmark	\\
腎	&	kidneys	&	B	&	\cmark	\\
臼齒	&	molar	&	M	&	\cmark	\\
花花公子	&	playboy	&	M	&	\cmark	\\
花花公子	&	dude	&	M	&	\cmark	\\
萎縮	&	atrophy	&	B	&	\cmark	\\
萎縮	&	shrinking	&	B	&	\cmark	\\
蒸	&	steamed	&	B	&	\cmark	\\
蒸	&	steaming	&	B	&	\cmark	\\
蓬	&	pont	&	M	&	\xmark	\\
虛擬	&	virtual	&	B	&	\cmark	\\
蠶	&	silkworm	&	M	&	\cmark	\\
蠶	&	silkworms	&	G	&	\cmark	\\
行為	&	behavioral	&	G	&	\cmark	\\
行為	&	behavior	&	B	&	\cmark	\\
行為	&	conduct	&	M	&	\cmark	\\
行為	&	behaviors	&	B	&	\cmark	\\
行為	&	behaviour	&	B	&	\cmark	\\
裝潢	&	decorating	&	M	&	\cmark	\\
要麼	&	either	&	B	&	\cmark	\\
觀測	&	observing	&	M	&	\cmark	\\
觀測	&	observations	&	G	&	\cmark	\\
觀測	&	observation	&	B	&	\cmark	\\
觀測	&	observed	&	G	&	\cmark	\\
觀測	&	observational	&	B	&	\cmark	\\
訂約	&	tendering	&	G	&	\xmark	\\
訂約	&	contracting	&	M	&	\cmark	\\
詩句	&	verses	&	B	&	\cmark	\\
詩句	&	verse	&	M	&	\cmark	\\
誤會	&	misunderstanding	&	B	&	\cmark	\\
誤會	&	mistaken	&	G	&	\cmark	\\
誤會	&	misunderstandings	&	G	&	\cmark	\\
負擔得起	&	affordable	&	B	&	\cmark	\\
貸方	&	lenders	&	B	&	\cmark	\\
貸方	&	lender	&	M	&	\cmark	\\
越野車	&	suv	&	M	&	\xmark	\\
跳舞	&	dancing	&	B	&	\cmark	\\
跳舞	&	dance	&	B	&	\cmark	\\
輸入	&	input	&	B	&	\cmark	\\
輸入	&	inputs	&	G	&	\cmark	\\
輸入	&	enter	&	M	&	\cmark	\\
轉型	&	transition	&	B	&	\cmark	\\
轉型	&	transformation	&	B	&	\cmark	\\
辯論	&	debate	&	B	&	\cmark	\\
辯論	&	debates	&	B	&	\cmark	\\
辯論	&	debating	&	M	&	\cmark	\\
農	&	nong	&	M	&	\xmark	\\
通過	&	via	&	B	&	\cmark	\\
通過	&	pass	&	M	&	\cmark	\\
通過	&	through	&	B	&	\cmark	\\
通過	&	passed	&	G	&	\cmark	\\
遊牧	&	nomadic	&	B	&	\cmark	\\
遊牧	&	nomad	&	M	&	\cmark	\\
過渡性	&	transitional	&	B	&	\cmark	\\
遺囑	&	wills	&	B	&	\cmark	\\
遺囑	&	testament	&	G	&	\cmark	\\
邊界	&	boundary	&	B	&	\cmark	\\
邊界	&	boundaries	&	B	&	\cmark	\\
邊界	&	border	&	B	&	\cmark	\\
邊界	&	borders	&	B	&	\cmark	\\
重複	&	duplicate	&	M	&	\cmark	\\
重複	&	repeated	&	G	&	\cmark	\\
重複	&	repeat	&	G	&	\cmark	\\
重複	&	repeats	&	G	&	\cmark	\\
重複	&	repeating	&	G	&	\cmark	\\
金庫	&	treasury	&	B	&	\cmark	\\
金庫	&	vault	&	M	&	\cmark	\\
金庫	&	vaults	&	G	&	\cmark	\\
開講	&	newsline	&	M	&	\xmark	\\
關稅	&	tariffs	&	B	&	\cmark	\\
關稅	&	customs	&	M	&	\cmark	\\
關稅	&	tariff	&	B	&	\cmark	\\
關節炎	&	arthritis	&	B	&	\cmark	\\
降低	&	reduce	&	G	&	\cmark	\\
降低	&	lowering	&	G	&	\cmark	\\
降低	&	reduced	&	G	&	\cmark	\\
降低	&	decrease	&	B	&	\cmark	\\
雀斑	&	freckles	&	M	&	\cmark	\\
雄心	&	ambitious	&	M	&	\cmark	\\
雄心	&	ambition	&	B	&	\cmark	\\
雄心	&	ambitions	&	B	&	\cmark	\\
雄蕊	&	stamens	&	B	&	\cmark	\\
雄蕊	&	petals	&	G	&	\xmark	\\
青色	&	cyan	&	M	&	\cmark	\\
鞋底	&	soles	&	M	&	\cmark	\\
鞋類	&	footwear	&	B	&	\cmark	\\
鞋類	&	shoes	&	G	&	\cmark	\\
飛行員	&	pilot	&	B	&	\cmark	\\
飛行員	&	pilots	&	G	&	\cmark	\\
食物	&	food	&	B	&	\cmark	\\
食物	&	foods	&	B	&	\cmark	\\
餘燼	&	embers	&	M	&	\cmark	\\
體	&	body	&	M	&	\xmark	\\
高貴	&	noble	&	B	&	\cmark	\\
魅力	&	charisma	&	B	&	\cmark	\\
魅力	&	charm	&	B	&	\cmark	\\
魅力	&	glamour	&	B	&	\xmark	\\
魅力	&	charismatic	&	G	&	\cmark	\\
鯊魚	&	sharks	&	B	&	\cmark	\\
鯊魚	&	shark	&	B	&	\cmark	\\
鵝	&	goose	&	B	&	\cmark	\\
鵝	&	geese	&	B	&	\cmark	\\
鹵素	&	halogens	&	G	&	\cmark	\\
鹵素	&	halogen	&	B	&	\cmark	\\
黑猩猩	&	chimps	&	M	&	\cmark	\\
黑猩猩	&	chimpanzee	&	B	&	\cmark	\\
黑猩猩	&	chimp	&	M	&	\cmark	\\
黑猩猩	&	chimpanzees	&	B	&	\cmark	\\
點頭	&	nod	&	B	&	\cmark	\\
齋戒	&	fasting	&	B	&	\cmark	\\
齋戒	&	ramadan	&	G	&	\xmark	\\
\end{xtabular}%
}

\end{CJK*}
\end{document}